\documentclass[journal]{IEEEtran}
\usepackage{cite,amsmath,amssymb,microtype}
\usepackage[pdftex]{graphicx}
\usepackage{threeparttable}
\usepackage[caption=false,font=footnotesize]{subfig}
\usepackage{xcolor}

\begin{document}

\title{Diversified Visual Attention Networks for Fine-Grained Object Classification}

\author{Bo~Zhao, Xiao~Wu$^{*}$,~\IEEEmembership{Member,~IEEE,}
        Jiashi~Feng, 
        Qiang Peng,
        and Shuicheng~Yan,~\IEEEmembership{Fellow,~IEEE}%
\thanks{
This work was supported in part by the National Natural Science Foundation of China (Grant Nos. 61373121, 61328205), Program for Sichuan Provincial Science Fund for Distinguished Young Scholars (Grant No. 13QNJJ0149), the Fundamental
Research Funds for the Central Universities, and China Scholarship Council (Grant No. 201507000032). \emph{Asterisk indicates corresponding author.}}
\thanks{Bo Zhao, Xiao Wu and Qiang Peng are with the School of Information Science and Technology, Southwest Jiaotong University, Chengdu, 610031, P.R. China, e-mail: zhaobo@my.swjtu.edu.cn, \{wuxiaohk, qpeng\}@home.swjtu.edu.cn.}%
\thanks{Jiashi Feng and Shuicheng Yan are with Department of Electrical and Computer Engineering, National University of Singapore, Singapore, e-mail: elefjia@nus.edu.sg, eleyans@nus.edu.sg.}%
}
\markboth{IEEE TRANSACTION ON MULTIMEDIA,~Vol.~X, No.~X, January~2017}
{Zhao \MakeLowercase{\textit{et al.}}: Diversified Visual Attention Network for Fine-Grained Object Classification}
\maketitle

\begin{abstract}

Fine-grained object classification attracts increasing attention in multimedia applications. However, it is a quite challenging problem due to the subtle inter-class difference and large intra-class variation. Recently, visual attention models have been applied to automatically localize the discriminative regions of an image for better capturing critical difference, which have demonstrated promising performance.
Unfortunately, without consideration of the diversity in attention process, most of existing attention models perform poorly in classifying fine-grained objects.
In this paper, we propose a diversified visual attention network (DVAN) to address the problem of fine-grained object classification, which substantially relieves the dependency on strongly-supervised information for learning to localize discriminative regions compared with attention-less models. More importantly, DVAN explicitly pursues the diversity of attention and is able to gather discriminative information to the maximal extent. Multiple attention canvases are generated to extract convolutional features for attention. An LSTM recurrent unit is employed to learn the attentiveness and discrimination of attention canvases. The proposed DVAN has the ability to attend the object from coarse to fine granularity, and a dynamic internal representation for classification is built up by incrementally combining the information from different locations and scales of the image. Extensive experiments conducted on CUB-2011, Stanford Dogs and Stanford Cars datasets have demonstrated that the proposed diversified visual attention network achieves competitive performance compared to the state-of-the-art approaches, without using any prior knowledge, user interaction or external resource in training and testing.
\end{abstract}


\section{Introduction}
\label{sec:introduction}
\IEEEPARstart{F}{ine-grained} object classification has attracted lots of attention from multimedia and computer vision communities, which aims to distinguish categories that are both visually and semantically very similar within a general category,\textit{ e.g.}, various species of birds~\cite{CUB_200_2011}, dogs~\cite{standford_dogs,Liu2012} and different classes of cars~\cite{standford_cars,Yang:2015ul}.
Fine-grained object classification is especially beneficial for multimedia information retrieval and content analysis. Examples include fine-grained image search~\cite{tmm04}, clothing retrieval and recommendation~\cite{r03,r06,r08}, food recognition~\cite{r09}, animal recognition~\cite{Liu2012,attention_categorization}, landmark classification~\cite{tmm07,wei2015hcp}, and so on.
Unfortunately, it is an extremely challenging task, because objects from similar subordinate categories may have marginal visual difference that is even difficult for humans to recognize. In addition, objects within the same subordinate category may present large appearance variations due to changes of scales, viewpoints, complex backgrounds and occlusions.

\begin{figure}[!t]
  \centering
  \subfloat[Great crested flycatcher]{\includegraphics[width=1\columnwidth]{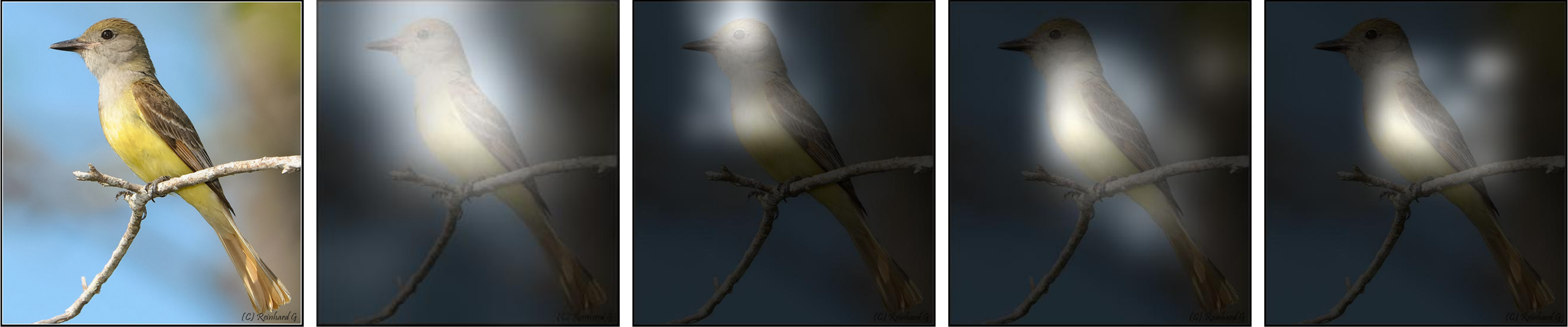}
  	\label{fig:diversified_attention_a}}
  \vspace{-0.1in}
  \hfil
  \subfloat[Yellow-breast chat]{\includegraphics[width=1\columnwidth]{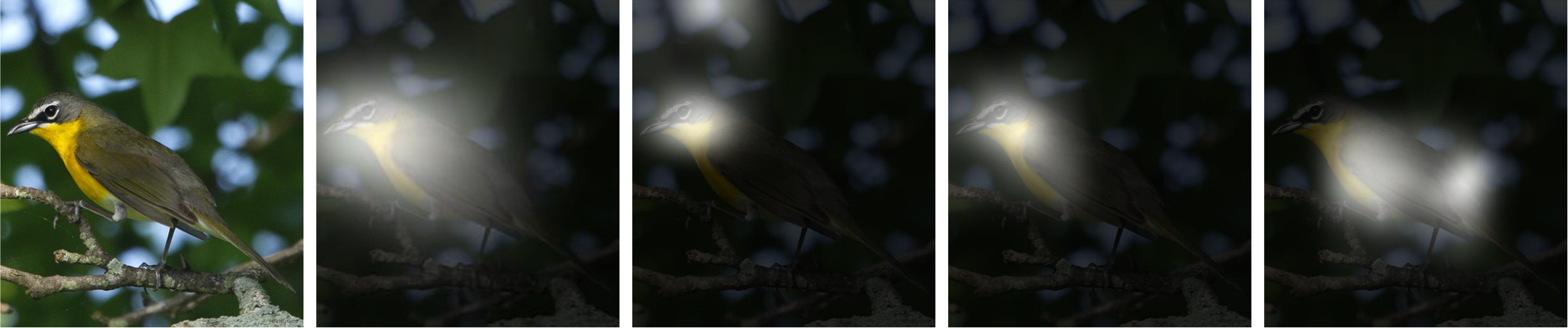}
  		\label{fig:diversified_attention_b}}
\caption{Two species of birds with similar appearance and their corresponding visual attention maps. It can be observed that the discriminative parts mainly locate at the eyes, breast and wings. The DVAN can automatically locate the main body of the bird and discover the discriminative parts of the two species of birds.
  }
  \label{fig:diversified_attention}
  \vspace{-0.2in}
\end{figure}

An important observation on fine-grained classification is that some local parts of objects usually play an important role in differentiating sub-categories. For instance, the heads of dogs are crucial for distinguishing many species of dogs. Motivated by this observation, most existing fine-grained classification approaches (\textit{e.g.},\cite{poof, kernal, pose_normalized, symbiotic_segmentation, codebook_free, part-based-rcnn, two_level_attention, fully_convolutional_attention}) first localize the foreground objects or object parts, and then extract discriminative features for classification. Region localization approaches~\cite{two_level_attention, fully_convolutional_attention} mainly employ unsupervised approaches to identify the possible object regions, while others (\textit{e.g.}, \cite{poof, kernal,pose_normalized, symbiotic_segmentation, codebook_free, part-based-rcnn}) alternatively use the available bounding box and/or part annotations. Unfortunately, these approaches suffer from some limitations. First, manually defined parts may not be optimal for the final classification task. Second, annotating parts is significantly more difficult than collecting image labels. Besides, manually cropping the objects and marking their parts are time consuming and labor intensive, which is not feasible for practical use. Third, the unsupervised object region proposal approaches generate a large number of proposals (up to several thousands), which is computationally expensive to process and classify the candidate regions.

Meanwhile, psychological research has also shown that humans tend to focus their attentions selectively on parts of the visual space, instead of processing the whole scene at once when recognizing objects, and different fixations over time are combined to build up an internal representation of the scene~\cite{ronald_2000, tmm01}. Such a visual attention mechanism is naturally applicable to fine-grained object classification. Therefore, many visual attention based networks have been proposed in recent years~\cite{action_recognition, attention_categorization, fully_convolutional_attention, multiple-object_visual-attention, two_level_attention, Xu:2015ut, Jaderberg:2015vo} and achieved promising results. However, it is difficult for existing visual attention models to find multiple visually discriminative regions at once. Besides, we also notice that the inter-class differences usually exist in small regions of an object, such as the beak or the legs for bird images. It is difficult for existing attention models to exactly localize them due to their small sizes. Therefore, zooming in these regions will be helpful for the attention model.

In this paper, we convert the problem of finding different attentive regions simultaneously to find them in multiple times. Owing to the capability of learning long-term dependencies in a recurrent manner, \emph{Long-Short-Term Memory (LSTM)}~\cite{Hochreiter:LSM} has been widely used in many deep neural networks for learning from experience to classify, process and predict time series. Therefore, LSTM is adopted to simulate the process of finding and learning multiple attentive regions of a fine-grained object.
Moreover, to solve the problem of small attention regions, a diversified attention canvas generation module is designed in our proposed DVAN. Given an image, multiple attention canvases are first generated with different locations and scales. Some of the canvases depict the whole object while others only contain certain local parts. An incremental representation for objects is dynamically built up by combining a coarse-grained global view and fine-grained diversified local parts. With this representation, the general picture and the local details of the objects can be captured to facilitate fine-grained object classification. Fig.~\ref{fig:diversified_attention} illustrates two species of birds with similar appearance. The main differences exist in the regions of eyes, breast and wings. Our diversified visual attention model can automatically discover and locate the subtle differences of these two species of birds on multiple attention canvases with location and scale variations. Another prominent merit of the proposed approach is that it does not need any bounding box or part information for both training and testing, which reduces the difficulty of large-scale fine-grained object classification.

The main contributions of this work are summarized as follows:
\begin{itemize}
  \item A diversified visual attention network (DVAN) is proposed for fine-grained object classification, which effectively and efficiently discovers and localizes the attentive objects as well as the discriminative parts of fine-grained images. To the best of our knowledge, this is the first work to exploit the diversity of computational attention for fine-grain object classification. More importantly, the approach does not need any prior knowledge or user interaction in training and testing.
  \item By combining a coarse-grained view and fine-grained diversified local parts, an incremental object representation is dynamically built up on the generated multiple attention canvases with various locations and scales, from which subtle differences can be accurately captured.
  \item The attention model is integrated with LSTM, in which recurrent neutrons are used to attend the sequential attention canvas from coarse to fine granularity.
  \item Experiments conducted on three benchmark datasets demonstrate that the proposed approach achieves competitive performance with state-of-the-art methods.
\end{itemize}

The rest of the paper is organized as follows. Related work is reviewed in Section~\ref{sec:related_work}. The proposed diversified visual attention network is elaborated in Section~\ref{sec:methods}. Experimental evaluation and analysis are presented in Section~\ref{sec:experiments}. Finally, we conclude this work in Section~\ref{sec:conclusion}.

\begin{figure*}[t]
  \centering
  \includegraphics[width=0.8\hsize]{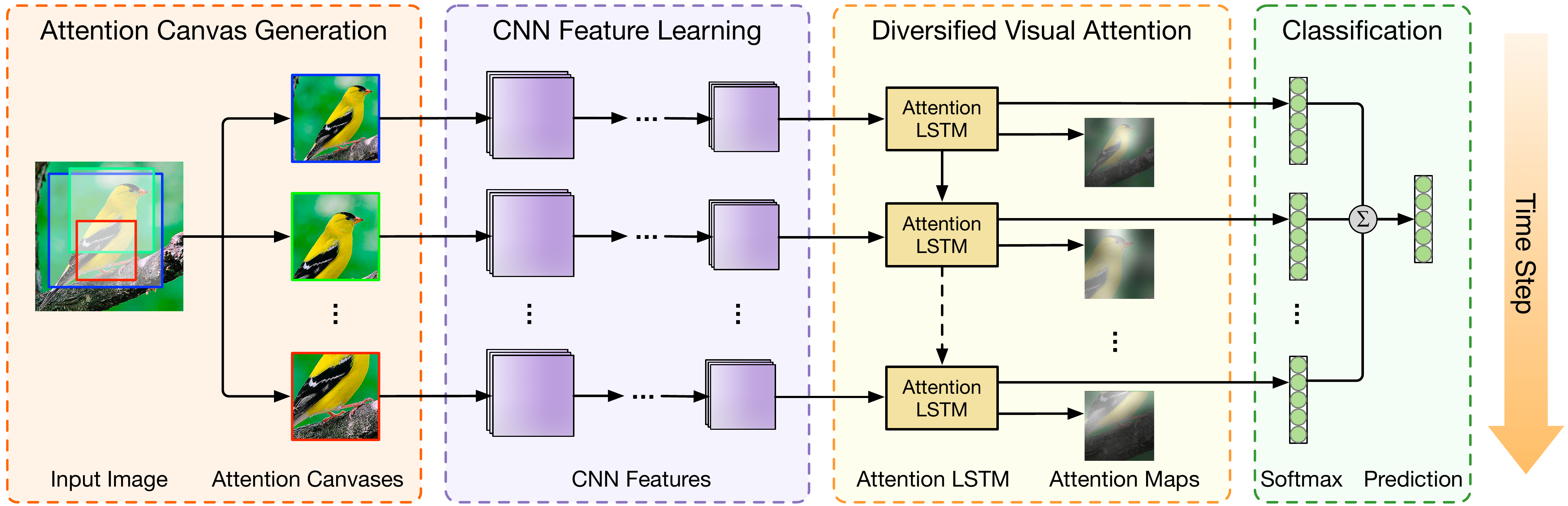}
  \vspace{-0.1in}
  \caption{The framework of the proposed Diversified Visual Attention Network (DVAN), which includes four components (from left to right): attention canvas generation, CNN feature learning, diversified visual attention, and classification. The attention canvas generation component prepares multiple canvases with different locations and scales, which are fed into the CNNs sequentially to learn the convolutional features. Based on these CNN features, the diversified visual attention model predicts the attention region of each attention canvas and dynamically pools the features. The attention model integrates the pooled attentive feature and predicts the object class at each time step. The final prediction is obtained by averaging the softmax values across all time steps.}
  \label{fig:framework}
  \vspace{-0.2in}
\end{figure*}

\section{Related Work}
\label{sec:related_work}

\subsection{Fine-Grained Image Classification}

Fine-grained image classification has been extensively studied in recent years. The approaches can be classified into the following four groups according to the use of additional information or human interaction.

\subsubsection{Discriminative Feature Representation}

Discriminative representation is very important for classification. Due to the success of deep learning in recent years, many methods~\cite{alexnet, decaf, Simonyan:2014ws} rely on the deep convolutional features, which achieve significantly better performance than conventional hand crafted features. A bi-linear architecture~\cite{bilinear} is proposed to compute the local pairwise feature interactions, by using two independent sub-convolutional neural networks. Motivated by the observation that a class is usually confused with a few other classes in multi-class classification, a CNN Tree~\cite{Wang:2015vo} is built to progressively learn find-grained features to distinguish a subset of classes by learning features only among these classes. Such features are more discriminative, compared to features learned for all classes. Similar to CNN Tree, Selective Regularized Subspace Learning~\cite{tmm06} conducts subspace learning only on confused classes with very similar viual appearance rather than the global subspace. The subset feature learning proposed in~\cite{Ge:2015tr} first clusters visually similar classes and then learns deep convolutional features specific for each subset.

\subsubsection{Alignment Approaches}

The alignment approaches find the discriminative parts automatically and align the parts. Unsupervised template learning~\cite{template} discovers the common geometric patterns of object parts and extracts the features within the aligned co-occurred patterns for fine-grained recognition. In alignment categorization\cite{alignment,Chai11}, images are first segmented and then objects are roughly aligned, from which features at different aligned parts are extracted for classification. The symbiotic model~\cite{symbiotic_segmentation} jointly performs segmentation and part localization, which help to update the results mutually. Finally, the features around each part are extracted for classification. By aligning detected keypoints to corresponding keypoints in a prototype image, lower-level features with normalized poses and higher-level features with unaligned images are integrated for bird species categorization in~\cite{improved_pose_normalized}.

\subsubsection{Part Localization-based Approaches}

These approaches usually localize important regions using a set of pre-defined parts, which can be either manually defined or automatically detected using data mining techniques. In~\cite{Liu2012}, the local appearance features of manually defined parts, such as face and eyes, are extracted and combined with global features for dog breed classification. The experimental results show that accurate part localization considerably increases classification performance. As an intermediate feature, Part-based One-vs-One Feature (POOF)~\cite{poof} is proposed to discriminate two classes based on the appearance feature of a particular part. Part-based R-CNNs~\cite{part-based-rcnn} learn the whole-object detector and part detectors respectively and predict a fine-grained category from a pose-normalized representation.

\subsubsection{Human-in-the-Loop Approaches}

Another branch is called human-in-the-loop methods~\cite{Branson2010, crowdsourcing_fine_grained, wah:similarity}, which need humans to interactively identify the most discriminative regions for fine-grained categorization. The main drawback is that it is not scalable for large scale image classification due to the need of human interaction.

\subsection{Visual Attention}
Existing visual attention models can be classified into soft and hard attention models. Soft attention models (\textit{e.g.},\cite{action_recognition, Jaderberg:2015vo}) predict the attention regions in a deterministic way. Therefore, it is differentiable and can be trained using back-propagation. Hard attention models (\textit{e.g.}, \cite{fully_convolutional_attention, Xu:2015ut, ram, multiple-object_visual-attention, attention_categorization}) predict the attention points of an image, which are stochastic. They are usually trained by reinforcement learning \cite{reinforcement} or maximizing an approximate variational lower bound. In general, soft attention models are more efficient than hard attention models, since hard attention models require sampling for training while soft attention models can be trained end-to-end.
Recurrent Attention Model \cite{ram} is proposed to learn the gaze strategies on cluttered digit classification tasks. It is further extended to multiple digit recognition~\cite{multiple-object_visual-attention} and less-constraint fine-grained categorization~\cite{attention_categorization}. The model can direct high resolution attention to the most discriminative regions without bounding box or part location. A two-level attention model is proposed in~\cite{two_level_attention} for fine-grained image classification. One bottom-up attention is used to detect candidate patches, and another top-down attention selects relevant patches and focuses on discriminative parts. The drawback is the two types of attention are independent and cannot be trained end-to-end, causing information loss and inefficiency.

\section{Diversified Visual Attention Networks}
\label{sec:methods}
In this section, the overall architecture of the proposed DVAN model is first introduced in Section \ref{sec:Arch}, followed by the visual attention model employed in DVAN (Section \ref{sec:VAM}). The core contributions, diversity promoting visual attention model and corresponding attention canvas generation approach, are then elaborated in Section \ref{sec:Diversity} and Section \ref{sec:Multiscale}, respectively, which effectively guarantee diversity and thus the information gain of the visual attention procedure.

\subsection{Overall Architecture}
\label{sec:Arch}
The architecture of the proposed DVAN model is described in Fig. \ref{fig:framework}, which includes four components: attention canvas generation, CNN feature learning, diversified visual attention and classification. DVAN first localizes several regions of the input image at different scales and takes them as the ``canvas'' for following visual attention. A convolutional neural network (i.e. VGG-16) is then adopted to learn convolutional features from each canvas of attention. To localize important parts or components of the object within each canvas, a diversified visual attention component is introduced to predict the attention maps, so that important locations within each canvas are highlighted and information gain across multiple attention canvases is maximized. Different from traditional attention models focusing on a single discriminative location, DVAN jointly identifies diverse locations with the help of a well designed diversity promoting loss function. According to the generated attention maps, the convolutional features will be dynamically pooled and accumulated into the diversified attention model. Meanwhile, the attention model will predict the object class at each time step. All the predictions will be averaged to obtain the final classification results.

\subsection{Visual Attention Model}
\label{sec:VAM}
\begin{figure}[!t]
  \centering
  \includegraphics[width=0.85\columnwidth]{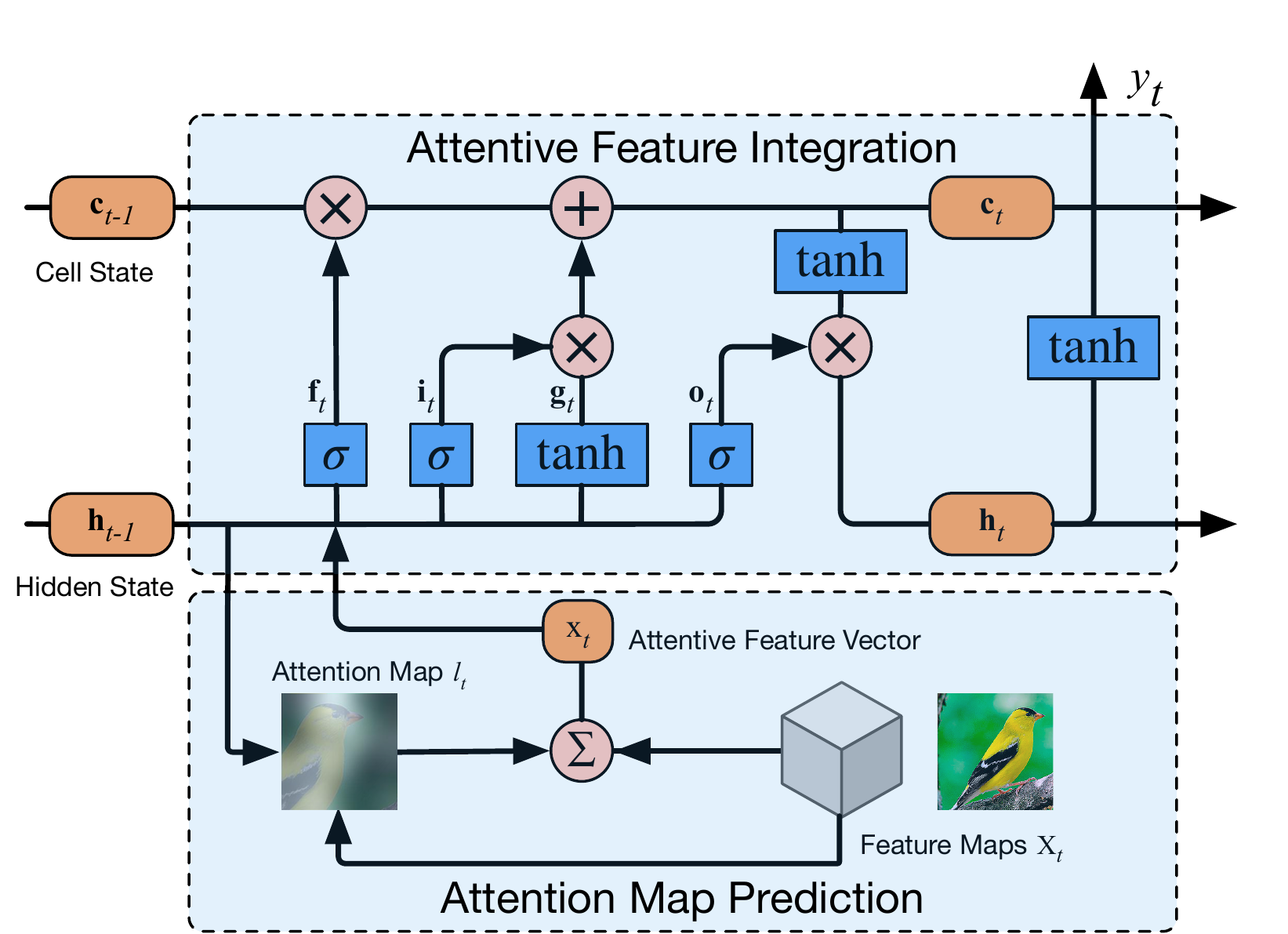}
  \vspace{-0.1in}  
  \caption{The visual attention model consists of two parts: attentive feature integration (top panel) and attention map prediction (bottom panel).}
  \label{fig:attention_lstm}
  \vspace{-0.2in}
\end{figure}

We will first introduce the visual attention model, which exploits the feature maps generated by the last convolution layer of the CNN. The adopted visual attention component in DVAN consists of two modules: attentive feature integration and attention map prediction, as demonstrated in the top and bottom panels of Fig. \ref{fig:attention_lstm}, respectively. The final prediction using the attentive features will be presented in the last part.

\subsubsection{Attention Feature Integration}
Since we want to convert the problem of simultaneously finding multiple attentive regions in an image into finding different attentive regions in multiple times, recurrent neural network is a preferable choice to solve this problem. More specifically, LSTM \cite{lstm} is chosen to perform the attentive feature integration in our attention model, due to its long short-term memory ability for modeling sequential data. To be self-contained, preliminary knowledge about LSTM model is first introduced. As shown in the top panel of Fig. \ref{fig:attention_lstm}, a typical LSTM unit consists of an input gate $\mathbf{i}_t$, a forget gate $\mathbf{f}_t$, an output gate $\mathbf{o}_t$ as well as a candidate cell state $\mathbf{g}_t$. The interaction between states and gates along time dimension is defined as follows:
\begin{align}
  \begin{pmatrix}
    \mathbf{i}_t\\ \mathbf{f}_t\\ \mathbf{o}_t\\ \mathbf{g}_t
  \end{pmatrix} &=
  \begin{pmatrix}
    \sigma \\ \sigma \\ \sigma \\ \text{tanh}
  \end{pmatrix}
  M
  \begin{pmatrix}
    \mathbf{h}_{t-1}\\ \mathbf{x}_t
  \end{pmatrix},\nonumber \\
  \mathbf{c}_t &= \mathbf{f}_t \odot \mathbf{c}_{t-1} + \mathbf{i}_t \odot \mathbf{g}_t,\\
  \mathbf{h}_t &= \mathbf{o}_t \odot \tanh\left(\mathbf{c}_t\right). \nonumber
\end{align}
Here $\mathbf{c}_t$ encodes the cell state, $\mathbf{h}_t$ encodes the hidden state, and $\mathbf{x}_t$ is the attentive feature generated by the attention map prediction module of our visual attention model.
The operator $ \odot $ represents element-wise multiplication. The pooled attentive feature vector $\mathbf{x}_t$  is then fed into LSTM unit and integrated with previous ones. The LSTM involves a transformation $M:\mathbb{R}^a \to \mathbb{R}^b$ that consists of $a\times b$ trainable parameters with $a=d+D$ and $b=4d$, where $d$ is the dimension of $\mathbf{i}_t$, $\mathbf{f}_t$, $\mathbf{o}_t$, $\mathbf{g}_t$, $\mathbf{c}_t$ and $\mathbf{h}_t$. Two activation functions, i.e., the \emph{sigmoid} function $\sigma$ and the function $\tanh$, are used in the attention model. The information flow and involved computation of three gates, as well as the cell and hidden states of LSTM unit, are illustrated in the top panel of Fig.~\ref{fig:attention_lstm}.

\subsubsection{Attention Map Prediction}
The process of attention map prediction is illustrated in the bottom panel of Fig.~\ref{fig:attention_lstm}. Let $\mathbf{X}_t$ be the feature maps generated by the last convolution layer of the CNN at time step $t$, which is represented as
\begin{equation}
\mathbf{X}_t = [\mathbf{X}_{t,1},\ldots,\mathbf{X}_{t,K},\mathbf{X}_{t,K+1},\ldots,\mathbf{X}_{t,K^2}],
\end{equation}
where $\mathbf{X}_{t,i}$ is the $i$-th slice of the feature maps $\mathbf{X}_t$ at time step $t$. Therefore, there are $K^2$ locations at the feature maps where DVAN can focus on. Each location is a feature vector where the dimensionality is the same as the number of the feature maps.

The feature maps $\mathbf{X}_t$ and hidden state of previous LSTM units $\mathbf{h}_{t-1}$ jointly determine the new attention map as follows:
\begin{equation}
  \label{eq:attention}
   l_{t,i}   =\frac{\exp(W_{h,i}^\top \mathbf{h}_{t-1} + W_{x,i}^\top \mathbf{X}_{t})}{\sum_{j=1}^{K^2}\exp(W_{h,j}^\top \mathbf{h}_{t-1} + W_{x,j}^\top \mathbf{X}_{t})},
             \forall i = 1, \ldots, K^2,
\end{equation}
where $W_{h,i}$ refers to weights of the connections from previous hidden state $\mathbf{h}_{t-1}$ to the $i$-th location of the spatial attention map. Similarly, $W_{x,i}$ denotes the weights from feature map $X_t$ to the $i$-th location of the map.

Then, the attentive feature $\mathbf{x}_t $, acting as the input of LSTM unit, is computed by the weighted summation over the feature map $\mathbf{X}_t$ based on the predicted attention map $l_t$:
\begin{equation}
  \mathbf{x}_t =\sum_{i=1}^{K^2}l_{t,i}\mathbf{X}_{t,i}.
  \label{eq:attention_pool}
\end{equation}

\subsubsection{Attention Model Initialization}
The cell and hidden states of LSTM are initialized using \emph{Multi-Layer Perceptron (MLP)}, and the average over all feature maps is used as the input of MLP:
\begin{align}
  \mathbf{c}_0&=f_{\text{init},c}\left(\frac{1}{TK^2}\sum^T_{t=1}\sum^{K^2}_{i=1}\mathbf{X}_{t,i}\right),\\
  \mathbf{h}_0&=f_{\text{init},h}\left(\frac{1}{TK^2}\sum^T_{t=1}\sum^{K^2}_{i=1}\mathbf{X}_{t,i}\right),
  \label{eq:h0}
\end{align}
where $f_{\text{init},c}$ and $f_{\text{init},h}$ are the functions implemented by two MLPs and $T$ is the total number of time steps of DVAN. These initial values are used to calculate the first attention map $l_1$, which determines the initial attentive feature $\mathbf{x}_1$.

\subsubsection{Classification}
The hidden state $\mathbf{h}_t$ of LSTM followed by a $\tanh$ activation function in the attentive feature integration module is used as the features for classification. The classification component is a fully convolutional layer followed by a softmax layer, which generates the probability of each category. Meanwhile, the hidden state $\mathbf{h}_t$ will also guide the next prediction of the attention map together with feature maps for next time step. In such a manner, new attentive features will be pooled according to the newly predicted attention maps. The states of LSTM will also be updated using the newly generated attentive features. The whole process will be recursively performed with the progress of time steps.
The final classification result is the average of the classification results across all time steps.

\subsection{Diversity Promoting Attention Models}
\label{sec:Diversity}

\begin{figure}[t]
	\centering
	\subfloat[Non-diversified visual attention maps.]{
		 \includegraphics[width=0.98\columnwidth]{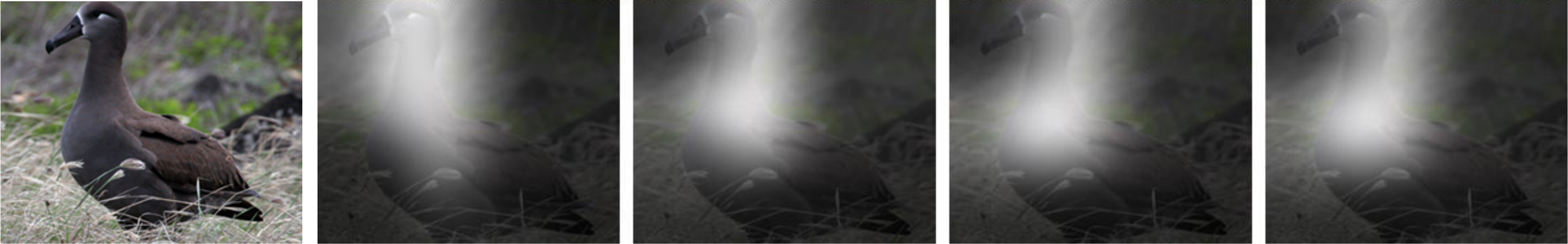}
		 \label{fig:diversify_comparison_a}
	}
    \vspace{-0.1in}
	\hfil
	\subfloat[Diversified visual attention maps.] {
		\includegraphics[width=0.98\columnwidth]{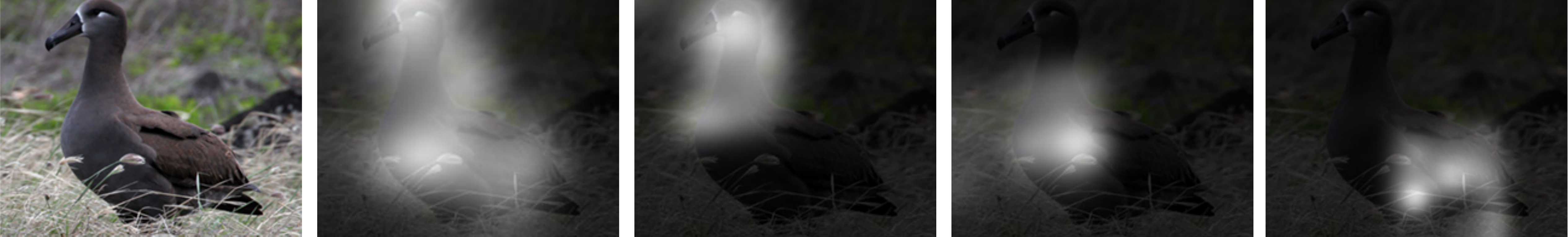}
		\label{fig:diversify_comparison_b}
		}
	\hfil
	\caption{The generated attention maps by vanilla (i.e., non-diversified) visual attention model vs. diversified visual attention on a bird image.}
	\label{fig:diversify_comparison}
	\vspace{-0.2in}
\end{figure}

The visual attention model introduced above has been demonstrated to be able to automatically localize discriminative regions for minimizing classification error in an end-to-end way~\cite{ram, action_recognition}. However, we observe that the attention maps generated at different time steps may be quite similar when the input image at each time step is the same. As a result, attention across different time steps does not gain additional information for better classification performance. To illustrate this issue more intuitively, the generated attention maps for a bird image at different time steps is visualized in Fig. \ref{fig:diversify_comparison}(a), in which vanilla visual attention model is used. From this figure, we can see that given the same input image, the attention model always focuses its attention on the head and neck of this bird across all time steps. Although the head and neck are discriminative for recognizing this species of birds from other genera or classes, they are insufficient to differentiate it from other visually similar species when the subtle differences lie in parts such as wings or tails. In addition, the beak or the eye area is too small, from which it is difficult to learn useful discriminative features for classification.

On the other hand, existing attention models only consider to minimize classification error during the attention process without concerning about information gain. It is defined as:
\begin{equation}
L_c=-\sum_{i=1}^{C}y_{i}\log\hat{y}_{i}
\end{equation}
where $y_i$ indicates whether the image belongs to class $i$. $C$ is the number of classes and $\hat y_i$ is the probability of class $i$.
Such strategy works well for classifying objects with significant difference. However, when the difference becomes quite subtle for fine-grained object classification, it is necessary to collect sufficient information from multiple small regions for making correct classification decision, which requires the attention process to be diverse. Therefore, in order to collect sufficient information for fine-grained object classification, a diversified attention model is novelly proposed in this work to capture diverse and discriminative regions.

To diversify the attention regions, the following diversity metric is proposed to compute the correlation between temporally adjacent attention maps:
\begin{equation}
  \label{eq:latt}
  L_\textit{div} = \frac{1}{T-1}\sum_{t=2}^T\sum_{i=1}^{K^2}l_{t-1,i} \cdot l_{t,i},
\end{equation}
where $l_{t,i}$ is the $i$-th attention value of the attention map after conducting softmax on $K^2$ locations at time step $t$. In general, $L_\textit{div}$ will get a large value if two neighboring attention maps are similar. Unfortunately, based on our empirical observation, solely minimizing the correlation measurement does not always provide sufficient diversity for attention maps.

To further enhance the diversity of attention, we force the visual attention model to check different locations of the image in the next time step. In order to achieve this goal, a ``hard'' constraint is imposed on the spatial support locations of the attention maps, which requires the overlapped proportion of temporal neighbouring attention regions to be smaller than a threshold, so that the attention regions can be shifted to different locations in neighbouring time steps.

The constrain is defined as follows,
\begin{equation}
\label{eq:supp}
\frac{\mathrm{Supp}[l_{t-1}] \cap \mathrm{Supp}[l_{t}]}{N} < \beta,\quad \forall t = 2,\ldots, T.
\end{equation}
where $\mathrm{Supp}[l_t]$ is the support region on the original image (i.e., the attention canvas), which is used to localize the attentive region. $N$ is the number of pixels of the original image. $\beta$ is a given threshold.

The final loss function combines the classification loss and diversity measure, along with hard constraint on the attention canvas, which is defined as:
\begin{align}
  L &=-\sum_{t=1}^T\sum_{i=1}^{C}y_{t,i}\log\hat{y}_{t,i} + \lambda L_{\textit{div}}, \label{eq:lambda}\\
  \text{s.t.} & \quad \frac{\mathrm{Supp}[l_{t-1}] \cap \mathrm{Supp}[l_{t}]}{N} < \beta,\quad \forall t = 2,\ldots, T. \nonumber
\end{align}
where $y_{t,i}$ is a one-hot label vector of class probabilities at time step $t$, $T$ is the total number of time steps, and $\lambda$ is a coefficient to control the extent of the penalty, if two neighboring attention locations do not change much.

\subsection{Multi-scale Attention Canvas Generation}
\label{sec:Multiscale}

To meet the constraint required by Eqn.~(\ref{eq:supp}) and make the attention map diversified, an attention canvas generation method is designed to crop multiple attention canvases with different locations and scales from the original image. It provides diversified canvases for our DVAN for attention. Some of the attention canvases contain main regions of the object, while others only include enlarged local regions of the object. These diversified attention canvases provide abundant candidates for DVAN to select discriminative regions from whole image level and enlarged local regions.

\begin{figure}[!t]
  \centering
  \includegraphics[width=1\columnwidth]{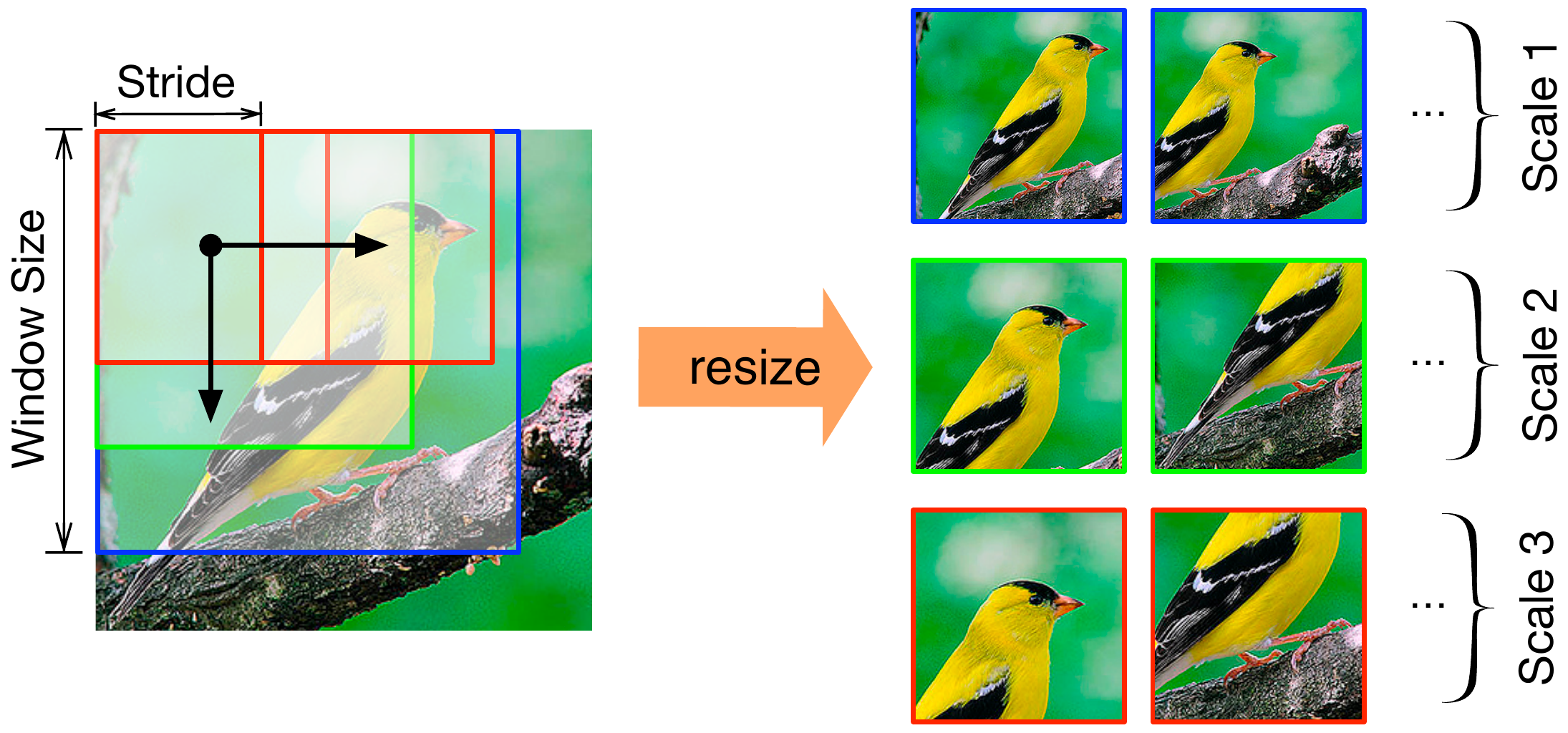}
  \vspace{-0.25in}
  \caption{Illustration on the generation of attention canvases using different window sizes and strides.}
  \label{fig:crop}
  \vspace{-0.2in}
\end{figure}

Since all attention canvases will be resized to the same size, the window size will represent the scale to enlarge. A large window size indicates a small scale for the resized attention canvas. Meanwhile, the resized attention canvas generated from small window size will have a large scale. Therefore, the local regions will be zoomed in and enlarged using a small window size. The stride determines the number of attention canvases to be generated. The attention canvas will be cropped according to the defined stride along the horizontal and vertical axes. More specifically, for a large square size, a large stride will generate a small number of canvases. On the contrary, given a smaller window size, more local regions will be cropped with a small stride. Finally, these attention canvases with various sizes will be normalized to the uniform size (e.g., $224 \times 224$ for VGG Net).
Fig.~\ref{fig:crop} illustrates the generation of attention canvases with different window sizes and strides.

Using this strategy, the generated attention canvases will cover most regions of the input image with different locations and scales. All the resized attention canvases will be organized into a sequence, so that the canvases with small scales will be placed before the canvases with large scales. In such a way, the visual attention model will first attend the main body of an object, and then local parts of the object will be further detected.

Fig.~\ref{fig:diversify_comparison}(b) demonstrates the diversified visual attention images after imposing the diversity penalty and multiple attention canvas generation. It can be observed that the input attention canvases are cropped at different locations and have various scales. These canvases form a sequence to be fed into DVAN. The visualized attention map of each image is diversified, from head to body, legs and tail, which is very reasonable for fine-grained object classification.

\section{Experiments}
\label{sec:experiments}
In this section, we evaluate the performance of DVAN model for fine-grained object classification. The benchmark datasets and implementation details of DVAN are first introduced. The model ablation studies are performed to investigate the contribution of each component. We compare DVAN with state-of-the-art methods, and the produced diversified attention maps are visualized in an intuitive way to demonstrate its superiority.


\subsection{Datasets}
\begin{figure}[!tb]
\centering
\subfloat[CUB Birds]{\includegraphics[width=1\columnwidth]{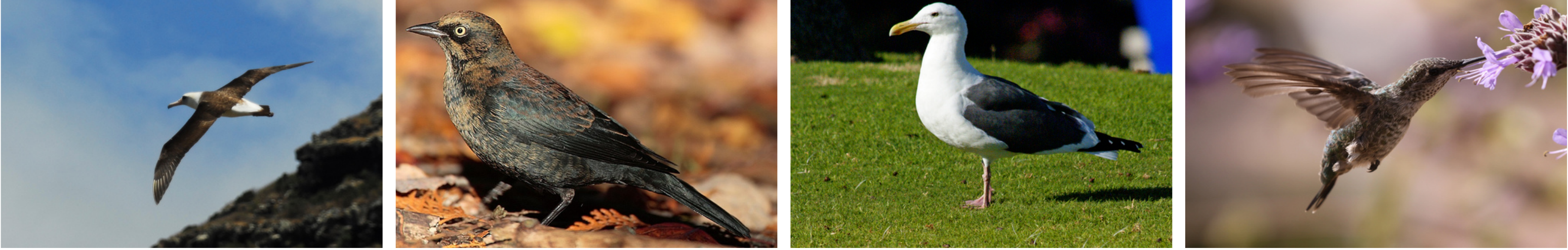}}
\vspace{-0.1in}
\hfil
\subfloat[Stanford Dogs]{\includegraphics[width=1\columnwidth]{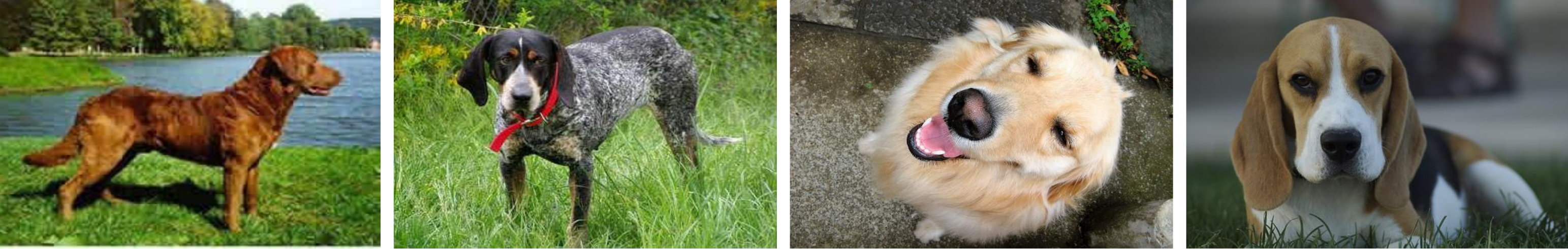}}
\vspace{-0.1in}
\hfil
\subfloat[Stanford Cars]{\includegraphics[width=1\columnwidth]{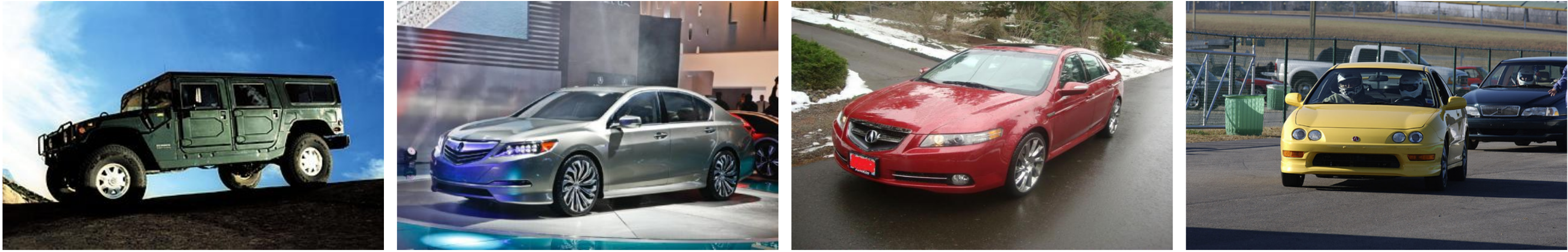}}
\hfil
\caption{Some examples from three fine-grained classification datasets used in the experiments. The images present cluttered background, which makes classification challenging.}
\label{fig:dogs_cars}
\vspace{-0.2in}
\end{figure}

We evaluate the performance of the proposed DVAN on three popular datasets for fine-grained object classification: Caltech-USCD Birds (CUB-200-2011)~\cite{CUB_200_2011}, Stanford Dogs \cite{standford_dogs} and Stanford Cars \cite{standford_cars}. Details about these three datasets are summarized in Table~\ref{tab:datasets}. Some representative images from these datasets are also shown in Fig.~\ref{fig:dogs_cars}. From these examples, we can see that the image contents are indeed complex, making the fine-grained classification rather challenging.

CUB-200-2011 dataset consists of 11,778 images from 200 bird categories, among which 5,994 images for training and 5,794 images for testing. It provides rich annotations, including image-level labels, object bounding boxes, attribute annotations and part landmarks.
Stanford Dogs dataset contains 20,580 images for 120 breeds of dogs. In each class, 100 images are used for training and around 71 images for testing. This dataset also provides tight bounding boxes surrounding the dogs of interest.
Stanford Cars dataset includes 16,185 images of 196 classes of cars. Class annotations are typically at the level of Year, Maker, Model. For example, 2012 Tesla Model S and 2012 BMW M3 coupe. In this dataset, most of the images have cluttered background. Therefore, visual attention is expected to be more effective for classifying them.

Note that although  bounding boxs or part-level annotations are available with these datasets,  our proposed model does not utilize any bounding boxs or part-level annotations throughout the experiments. This is one of the advantages of the proposed method.

\begin{table}[tp]
  \centering
  \caption{Dataset statistics and information on provided annotations. BBox stands for  bounding box, and Part  means  part-level annotation.}
  \vspace{-0.1in}  
  \label{tab:datasets}
  \begin{tabular}{l c c c c c}
    \hline\hline
    Dataset & \# Class & \# Training & \# Testing & BBox & Part \\
    \hline
    CUB-200-2011 \cite{CUB_200_2011}& 200 & 5,994 & 5,794 & $\surd$ & $\surd$\\
    Stanford Dogs \cite{standford_dogs}& 120 & 12,000 & 8,580 & $\surd$ & \\
    Stanford Cars \cite{standford_cars}& 196 & 8,144 & 8,041 & $\surd$ & \\
   \hline\hline
  \end{tabular}
\vspace{-0.2in}
\end{table}

\subsection{Implementation Details}
In this subsection, we will describe the training details of the proposed DVAN model.
All images are first normalized by resizing their short edges to 256 and keeping their aspect ratios. For attention canvas generation, three different window sizes are used, i.e., $224\times 224$, $168\times 168$ and $112 \times 112$ to generate canvases from the resized images. Accordingly, the stride is set as 32, 44 and 48, respectively, to generate the image canvases along x and y axes. Finally, we obtain $2 \times 2$, $3 \times 3$ and $4 \times 4$ attention canvases for these three window sizes. In addition, one central region of the image is also kept for each scale. Totally, we obtain 32 different attention canvases with three scales. All these attention canvases are resized to the same size, i.e., $224 \times 224$ for VGG-16 Networks. The sequence of these canvases is arranged in this way: the first 5 canvases is from scale of $224 \times 224$, followed by the 10 canvases from scale of $168 \times 168$, and the last 17 canvases is from scale of $112 \times 112$.

The popular VGG Net~\cite{Simonyan:2014ws} is deployed to extract the CNN features, which includes 13 convolutional layers and 3 fully connected layers. The output of the last convolution layer will be used as the input of the visual attention model. The dimensionality of the hidden and cell states of LSTM and the hidden layer of MLP to initialize LSTM are set to 512. A three-stage training approach is adopted for the sake of speed. First, we fine-tune the CNN model pre-trained on ImageNet to extract the basic convolutional feature maps for attention localization. Second, the generated basic convolutional feature maps are fixed, and the diversified attention model is trained separately. Finally, the whole network is updated to further improve the performance. The model is trained using \emph{stochastic gradient descent} (SGD) for 50 epochs at each stage with learning rate of 0.001. Our implementation is based on \emph{Lasagne}\footnote{https://github.com/Lasagne}, which is a light-weight library to build and train neural networks in \emph{Theano}\footnote{https://github.com/Theano}. The sequence of attention canvases with same scale are randomly selected for training, and the fixed order of all canvases of three scales are adopted for testing.

\subsection{Model Ablation Studies}
In this subsection, we evaluate the effect of attention mechanism and different parameter settings for the performance of DVAN.

\begin{table}[!bt]
  \centering
  \vspace{-0.1in}
  \caption{Effect of multiple attention canvases and different pooling strategies on CUB-200-2011 dataset}
  \label{tab:comparison_pool_cub}
  \begin{tabular}{lc}
    \hline\hline
    Model                & Acc (\%)\\
    \hline
    VGG-16               & 72.1  \\
    VGG-16-multi-canvas & 74.5  \\
    \hline
    DVAN-Avg  & 76.8  \\
    DVAN-Max      & 76.2  \\
    DVAN       & \textbf{79.0}\\
    \hline\hline
  \end{tabular}
  \vspace{-0.2in}
\end{table}

\subsubsection{Effect of Attention Mechanism}
To evaluate the impact of the attention mechanism of the DVAN, we train a VGG-16 model using multiple attention canvases as a baseline. It can be treated as replacing the diversified visual attention component in DVAN with two fully connected layers with size of 4096. The final prediction is also the average of the softmax values of all attention canvases as well as DVAN. The VGG-16 trained with single image is chosen as another baseline. The results of VGG-16 and VGG-16 with multiple attention canvases are listed in the upper part of Table \ref{tab:comparison_pool_cub}. The accuracy of VGG-16 trained by single image is 72.1\%. By augmenting the training data with multiple attention canvases, VGG-16-multi-canvas improves the accuracy to 74.5\%. After adopting visual attention, DVAN can significantly improve the performance to 79.0\%.

\subsubsection{Effect of Different Pooling Methods}
The visual attention component in DVAN provides a dynamic pooling strategy to highlight the important and discriminative regions of the image. To prove its effectiveness, we compare it with DVAN using other pooling strategies, such as max pooling and average pooling.
The average pooling DVAN (DVAN-Avg) and max pooling DVAN (DVAN-Max) have the identical architecture as the proposed DVAN model, except that they do not predict the attention map on the feature maps. Instead of computing the attentive feature vector, the feature maps are average pooled or max pooled as the input feature for the LSTM in visual attention model.

The performance comparison is listed in Table~\ref{tab:comparison_pool_cub}. Both DVAN-Avg and DVAN-Max have better performance than VGG-16-multi-canvas, demonstrating the superiority of LSTM to model sequential data. The classification accuracy of DVAN-Avg is 76.8\%, which is slightly better than DVAN-Max. Without dynamically weighted pooling, DVAN-Avg will be interfered by the features of trivial regions. On the other hand, retaining only the features with max response, DVAN\_Max will lose more important information compared with averaged pooling. The proposed DVAN model can learn the discriminative regions from feature maps, and dynamically pool them according to the attention probabilities. Its performance reaches 79.0\%, outperforming both DVAN-Avg and DVAN-Max.

\subsubsection{Effect of Parameter $\lambda$}
\begin{figure}[!t]
\centering
\subfloat[Attention maps with $\lambda=0$]
{\includegraphics[width=1\columnwidth]{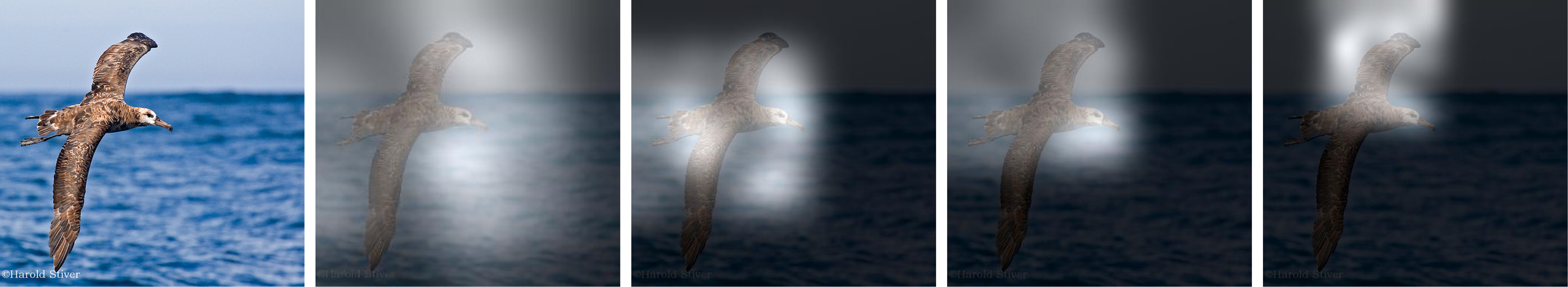}}
\vspace{-0.1in}
\hfil
\subfloat[Attention maps with $\lambda=1$]
{\includegraphics[width=1\columnwidth]{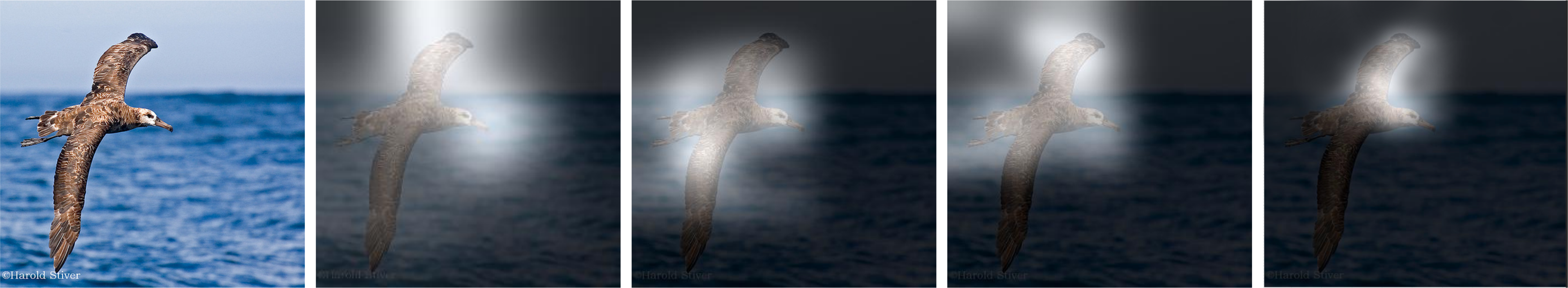}}
\vspace{-0.1in}
\hfil
\subfloat[Attention maps with $\lambda=10$]
{\includegraphics[width=1\columnwidth]{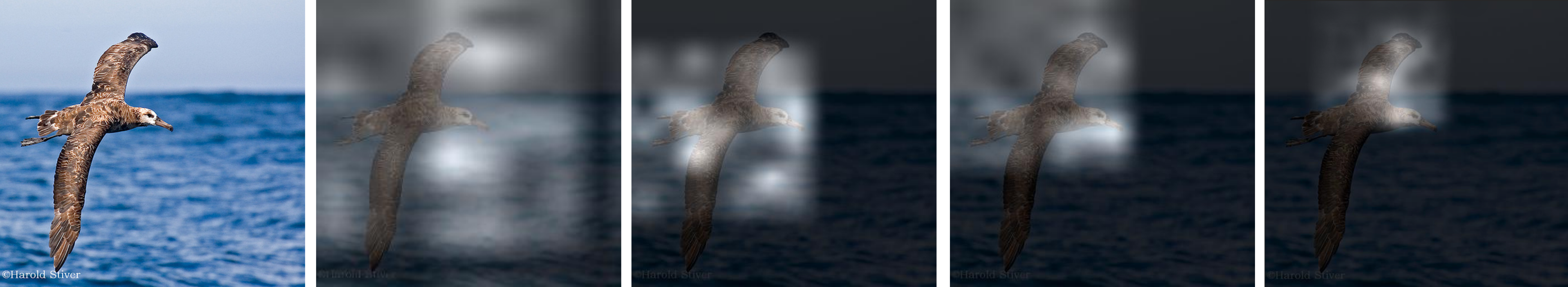}}
\hfil
\caption{The original image and its corresponding attention maps generated by different values of parameter $\lambda$, (a) $\lambda=0$ (b) $\lambda=1$ (c) $\lambda=10$.}
\label{fig:lambda}
\end{figure}

\begin{figure}[t]
  \centering
  \includegraphics[width=.7\columnwidth]{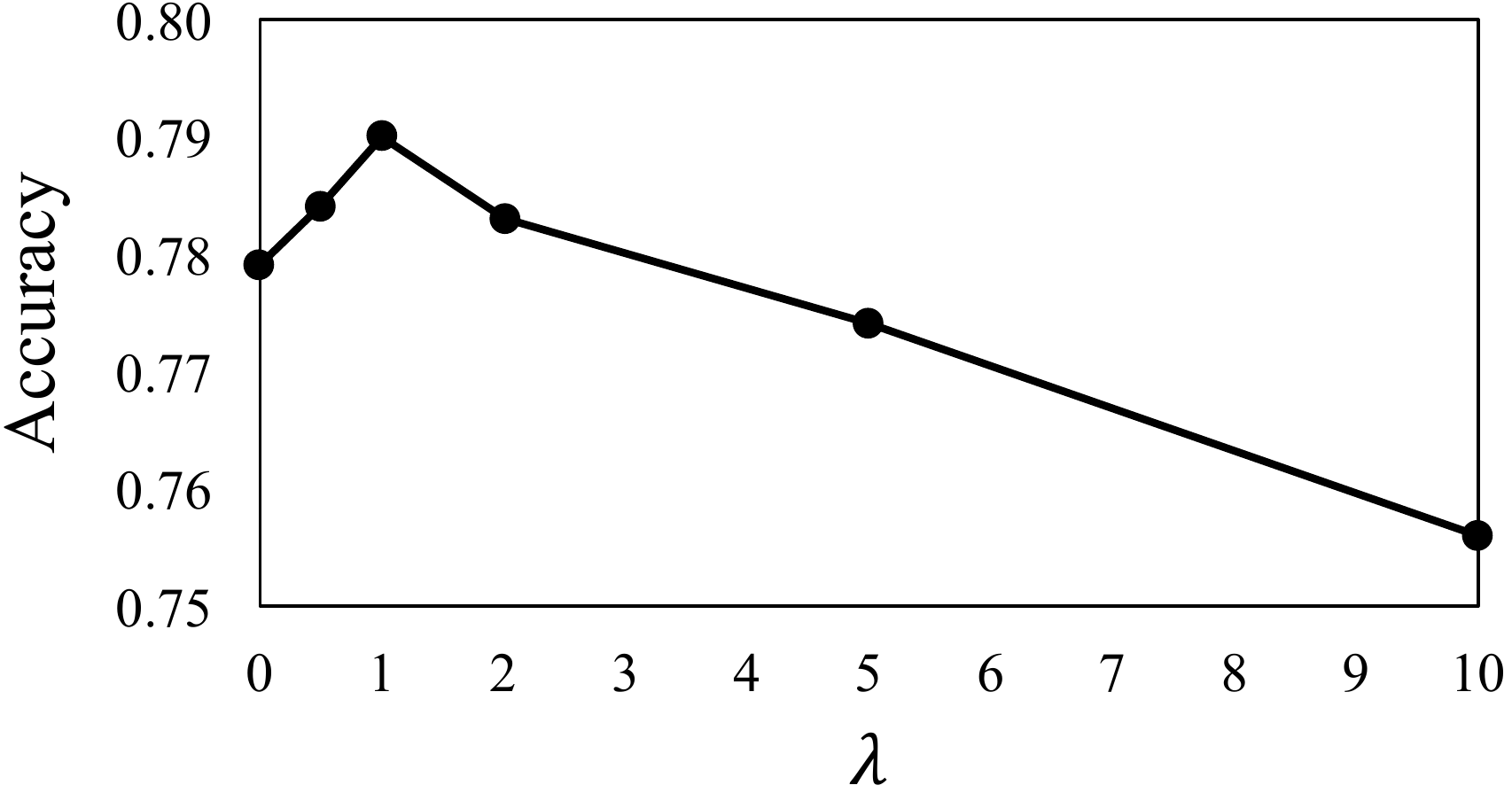}
  \vspace{-0.1in}
 \caption{Performance of DVAN on CUB-200-2011 dataset with different $\lambda$.}
  \label{fig:comparison_lambda}
    \vspace{-0.2in}
\end{figure}

$\lambda$ is a parameter used in Eqn. (\ref{eq:lambda}) to control the diversity of the neighboring attention maps. The effect of parameter $\lambda$ for the performance of DVAN model is demonstrated in Fig.~\ref{fig:comparison_lambda}. When $\lambda = 0$, it does not require the successive attention maps to be diversified. On the contrary, a larger value strongly enforces the DVAN to attend different regions of the attention canvas. The performance is first improved and then dropped after $\lambda$ becomes larger. Our attention model better observes discriminative regions as $\lambda$ increases. Unfortunately, strong diversity constraint is imposed for attending the object as continuously increasing $\lambda$. More attention fixations are scattered in the images, leading to the loss of important information.
It achieves the best performance (79.0\%) when $\lambda$ is equal to 1, which will be used as the default setting in our later experiments.

The visualized attention maps with three different settings are illustrated in Fig.~\ref{fig:lambda}. It can be observed that the major areas of the image are attended in Fig.~\ref{fig:lambda}(a) when $\lambda=0$, but a large portion of backgrounds are falsely involved. The mistakenly attended background regions will reduce the performance of classification. As demonstrated in Fig.~\ref{fig:lambda}(b), the attention regions more accurately focus on the torso or the local parts of the bird when $\lambda=1$. However, when $\lambda$ becomes larger, it can be seen from Fig.~\ref{fig:lambda}(c) that the attention regions are unevenly scattered and they are forcibly shifted to different locations to meet the strong constraint. The discriminative regions are easily ignored while more irrelevant regions are contained, causing performance degradation.

\subsubsection{Performance of Different Scales}
Multi-scale attention canvases at different locations contribute to the diversity of attention maps, so that our model can observe the object from coarse to fine granularity. To reveal the performance with different scales, the experiments with different combinations of scales are conducted. One-scale attention only uses the attention canvases generated by window size of $224 \times 224$. Two-scale attention model adopts window sizes of $224 \times 224$ and $168 \times 168$, while three-scale attention model attends the canvases with three different window sizes, $224 \times 224$, $168\times 168$ and $112\times 112$. For a good trade-off among training efficiency, GPU memory cost and performance, experiments with more scales are not conducted. Note that all generated attention canvases will be resized to the same dimensionality (i.e., $224\times 224$) for training and testing. 

The performance comparison on different datasets is summarized in Table \ref{tab:comparison_cub_1}. The fine-tuned VGG-16 model is chosen as the baseline. Using the original image as the input, the performance for VGG-16 model is the worst. The general trend is that the performance is significantly improved, when more scales of canvases are involved in the diversified visual attention model. The accuracy reaches 79.0\% and 81.5\% for CUB-200-2011 and Stanford Dogs, respectively, when three-scale attention model is adopted. It proves that the diversified visual attention model can correctly detect most discriminative regions  with different scales, which achieves the best performance. In general, the generated attention canvases with large window sizes tend to capture the whole object, while detailed regions will be attended with small window sizes. An interesting phenomenon is that the accuracy of two-scale attention is better than three-scale attention for Stanford Cars dataset. It indicates that the attention canvases generated by a small window size cannot always provide additional information for fine-grained object classification. For example, if the attention canvases only contain the tires or windshields of a car, they are not capable of distinguishing different models, since the wheels or windshields are visually similar across different types of cars.

\begin{table}[!t]
  \centering
  \caption{Performance comparison with different scales}
  \label{tab:comparison_cub_1}
  \begin{tabular}{lccc}
    \hline\hline
    Model                            & CUB-200-2011 & Stanford Dogs & Stanford Cars\\
    \hline
    VGG-16                           & 72.1     & 76.7  & 83.8\\
    One-scale Attention              & 73.4     & 78.0  & 85.1\\
    Two-scale Attention              & 77.0     & 79.4  & \textbf{87.1}\\
    Three-scale Attention            & \textbf{79.0} & \textbf{81.5} & 86.4 \\
    \hline\hline
  \end{tabular}
  \vspace{-0.2in}
\end{table}

\begin{table}[!t]
  \centering
  \caption{Effect of CNN Feature Learning on CUB-200-2011 dataset}
  \vspace{-0.1in}
  \label{tab:pool5}
  \begin{tabular}{lc}
    \hline\hline
    Model                & Acc (\%)\\
    \hline
    VGG-16 (pool5)              & 65.9\\
    VGG-16-multi-canvas (pool5) & 66.5\\
    DVAN (pool5)                & \textbf{68.2}\\
    \hline\hline
  \end{tabular}
  \vspace{-0.2in}
\end{table}

\subsubsection{Effect of CNN Feature Learning} Since the gradients of the diversified visual attention module can be back-propagated to the CNN feature learning module, DVAN is end-to-end trainable and learns better features. To verify this, we extract the \textit{pool5} features of the trained DVAN, VGG-16 and VGG-16-multi-canvas. We train SVM classifiers on these features and then use them to classify test images. The results are reported in Table~\ref{tab:pool5}. It can be seen that the accuracy of DVAN (pool5) is 68.2\%, which is better than the one trained by the pool5 features of VGG-16 or VGG-16-multi-canvas. The results demonstrate that the learned convolutional features of DVAN are superior over the ones of VGG-16 and VGG-16-multi-canvas.
\subsection{Comparison with State-of-the-Art Methods}
\subsubsection{Performance on CUB-200-2011}
\begin{table}[!t]
  \centering
  \caption{Comparison with state-of-the-art methods on CUB-200-2011}
  \vspace{-0.1in}
  \label{tab:comparison_cub_2}
  \begin{tabular}{lccc}
    \hline\hline
    Model                & BBox & Annot. & Acc (\%)\\
    \hline
    POOF~\cite{poof}     & $\surd$   & $\surd$   & 73.3\\
    Pose Normalization~\cite{pose_normalized}  & $\surd$   & $\surd$   & 75.7\\
    Part-based RCNN~\cite{part-based-rcnn}          & $\surd$   & $\surd$   & 76.4\\
    Two-level Attention~\cite{two_level_attention}& & & 77.9\\
    DVAN       &           &           & 79.0\\
    NAC~\cite{nac} & & & 81.0\\
    ST-CNN~\cite{Jaderberg:2015vo} & & & 82.3\\
    Bilinear CNN~\cite{bilinear} & & & 84.1\\
    PD~\cite{Zhang_2016_CVPR} & & &\textbf{84.5}\\
    \hline\hline
  \end{tabular}
  \vspace{-0.2in}
\end{table}
We compare the proposed framework with state-of-the-art methods on CUB-200-2011 dataset, which is listed in Table~\ref{tab:comparison_cub_2}. POOF~\cite{poof} learns a large set of discriminative intermediate-level features, specific for a particular domain and a set of parts. The bounding box information and part annotation are used for part feature extraction and comparison. The accuracy is 73.3\%. Pose Normalization \cite{pose_normalized} computes local image features with pose estimation and normalization. Low-level pose normalized features and high-level features are combined for final classification, which improves the performance to 75.7\%. The idea of part-based RCNN~\cite{part-based-rcnn} is similar to pose normalization, in which the whole object and part detectors are learned with geometric constraints of parts and the object. A pose-normalized representation is formed for classification. Beneficial from more accurate part detectors, the accuracy of part-based RCNN is 76.4\%. Unlike aforementioned methods using bounding box information or part annotation, Two-level Attention~\cite{two_level_attention} combines object-level attention and part-level attention to find the interested parts of an object, which selects image patches relevant to the target object and uses different CNN filters as the part detectors. The overall precision is 77.9\%. Unlike two-level attention using two independent components and involving proposals generated by other algorithms, DVAN adopts a uniformed mechanism to perform the object-level and part-level attention detection using LSTM. The accuracy of our proposed DVAN is 79.0\%, which is better than the Two-level Attention method and aforementioned methods using bounding box information and part annotations.
By using a deeper Inception architecture with batch normalization~\cite{bn} and actively spatial transforming feature maps, ST-CNN~\cite{Jaderberg:2015vo} achieves better accuracy (82.3\%). Bilinear CNN~\cite{bilinear} utilizes two CNNs (M-Net~\cite{Chatfield14} and D-Net~\cite{Simonyan:2014ws}) to model local pairwise feature interactions in a translationally invariant manner. At the cost of additional feature extraction, Bilinear CNN~\cite{bilinear} obtains the performance of 84.1\%.
Different from our attention model which implicitly finds the discriminative parts of the birds, NAC~\cite{nac} and PD~\cite{Zhang_2016_CVPR} explicitly use specific channels of a CNN as part detectors. Selective Search~\cite{selective} is also used in these two methods to better locate the object or its parts. Based on the detected parts, NAC learns the part models and extract features at object parts for classification. PD encodes the part filter responses using Spatial Weighted Fisher Vector and pools them into the final image representation. The more complicated models adopted in NAC and PD make their performances better than ours.

\subsubsection{Performance on Stanford Dogs}
\begin{table}[!t]
  \centering
  \caption{Comparison with state-of-the-art methods on Stanford Dogs}
  \vspace{-0.1in}
  \label{tab:comparison_dogs_2}
  \begin{tabular}{lccc}
    \hline\hline
    Model & BBox & Web Data & Acc(\%)\\
    \hline
    Alignment localization \cite{alignment} & $\surd$ &   & 50.1\\
    Unsupervised grid alignments~\cite{Gavves2015}  & $\surd$ &   & 56.7\\
    Glimpse attention~\cite{attention_categorization} & & & 76.8\\
    Noisy data CNN~\cite{noisy}&& $\surd$  & \textbf{82.6}\\
    DVAN                       & & & 81.5\\
    \hline\hline
  \end{tabular}
  \vspace{-0.2in}
\end{table}
The performance comparison with state-of-the-art approaches on Stanford Dogs dataset is listed in Table~\ref{tab:comparison_dogs_2}. Unlike the works relying on detectors for specific object parts, alignment localization~\cite{alignment} identifies distinctive regions by roughly aligning the objects using their implicit super-class shapes. Although the pre-given bounding boxes are used during training and testing, the performance is very poor, only 50.1\%. It proves that the proposed unsupervised alignment approach is not well adaptive for this dataset. Without any spatial supervision such as bounding boxes, glimpse attention \cite{attention_categorization} directs high resolution attention to the most discriminative regions, with reinforcement learning-based recurrent attention models. Its recognition accuracy is significantly boosted, reaching 76.8\%. By exploiting additional web data from Google Image Search to expand training datasets, the noisy data CNN~\cite{noisy} achieves the best performance, 82.6\% accuracy. Without using the bounding box information and additional Web data augmentation, the performance of our approach reaches 81.5\%, which is much better than the approaches using bounding box information as well as glimpse attention algorithm. The noisy data CNN obtains slightly better accuracy at the expense of additional web data.

\subsubsection{Performance on Stanford Cars}
\begin{table}[!t]
  \centering
  \caption{Comparison with state-of-the-art methods on Stanford Cars}
  \vspace{-0.1in}  
  \label{tab:comparison_cars_2}
  \begin{tabular}{lcc}
    \hline\hline
    Model                   & BBox Info & Acc (\%)\\
    \hline
    ELLF~\cite{ellf}        & $\surd$ & 73.9\\
    Symbiotic segmentation~\cite{symbiotic_segmentation}& $\surd$ & 78.0\\
    Fisher vector~\cite{revisit_fv}    & $\surd$   & 82.7\\
    R-CNN~\cite{rich_feature}          & $\surd$   & \textbf{88.4}\\
    DVAN                               &           & 87.1\\
    \hline\hline
  \end{tabular}
  \vspace{-0.2in}
\end{table}
The performance comparison on Stanford Cars is summarized in Table \ref{tab:comparison_cars_2}. ELLF~\cite{ellf} uses the stacked CNN features which are pooled in the detected part regions as the final representation, obtaining 73.9\% classification accuracy. Symbiotic segmentation~\cite{symbiotic_segmentation} improves the accuracy to 78.0\% by deploying a symbiotic part localization and segmentation model. Fisher vector \cite{revisit_fv} further improves the performance to 82.7\%, proving the effectiveness of orderless pooling methods. With the assistance of bounding boxes, R-CNN~\cite{rich_feature} applies high-capacity convolutional neural networks to bottom-up region proposals in order to localize and segment objects, yielding the best accuracy of 88.4\%. All these methods resort to bounding box information for better object localization. Different from these approaches, DVAN locates the discriminative parts automatically without the requirement of any auxiliary information, achieving competitive performance compared to R-CNN (87.1\% vs. 88.4\%).

\subsection{Diversified Attention Visualization}
\begin{figure}[!t]
  \centering
  \subfloat[Black Footed Albatross from CUB-200-1011]{\includegraphics[width=0.9\hsize]{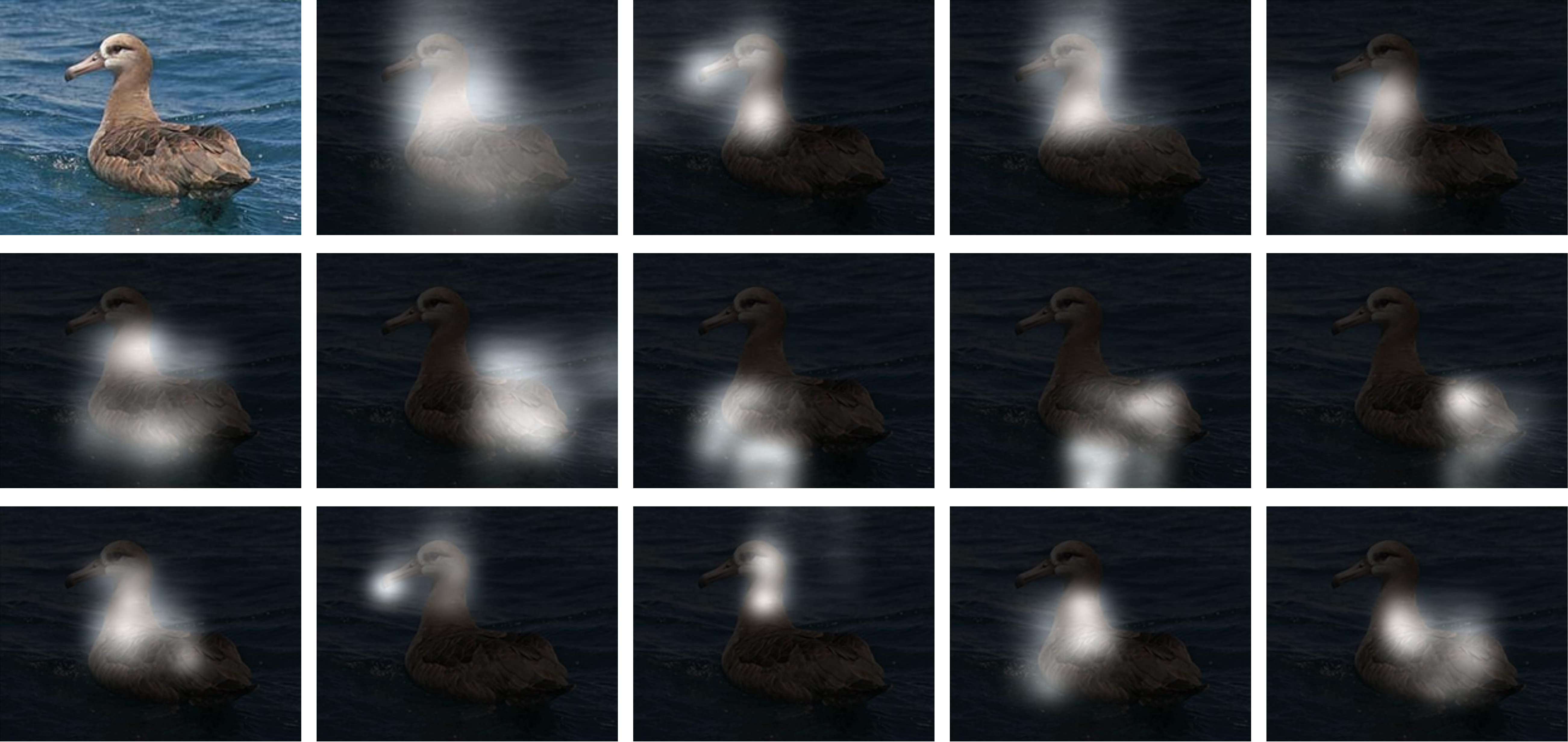}}
  \vspace{-0.1in}
  \hfil
  \subfloat[Blenheim Spaniel from Stanford Dogs]{\includegraphics[width=0.9\hsize]{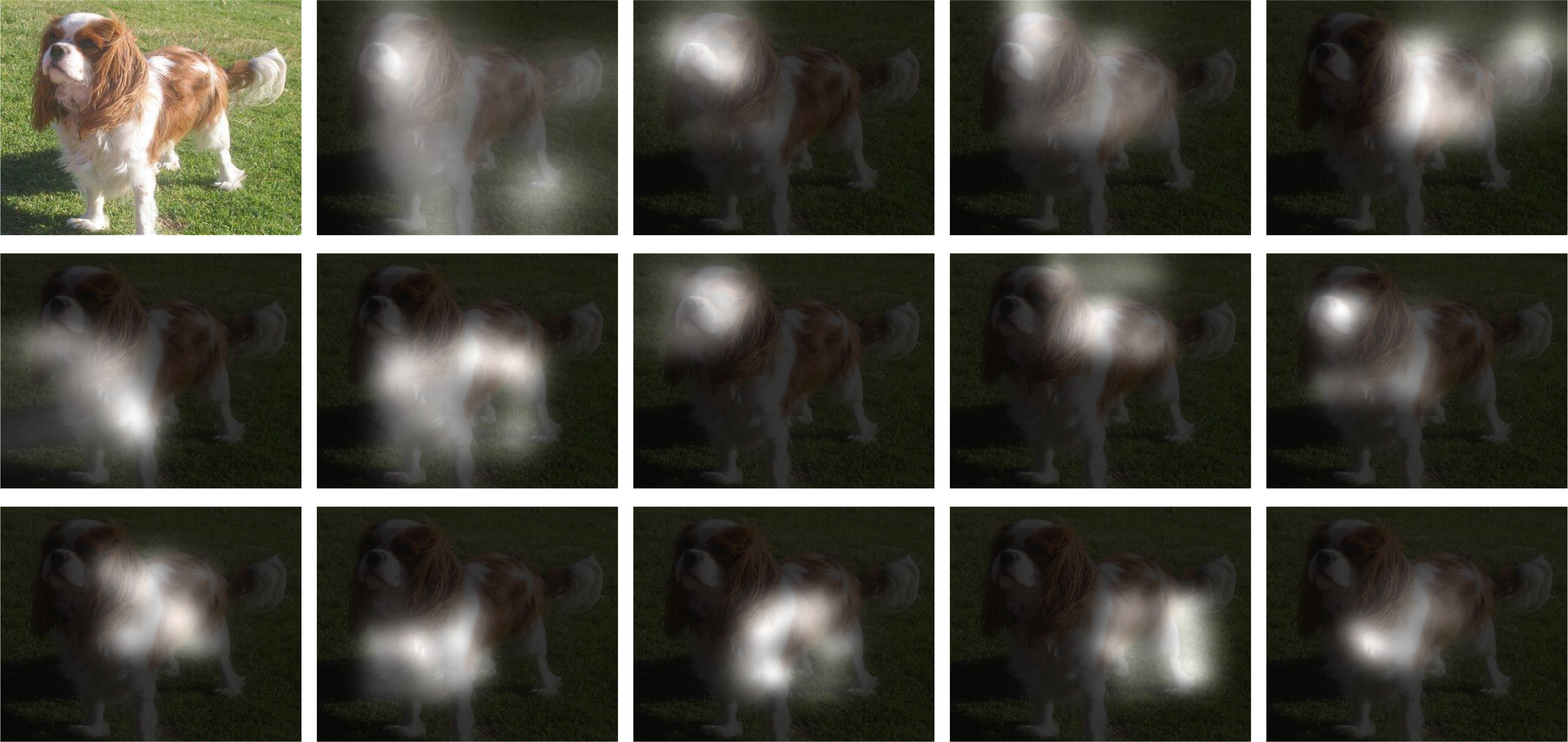}}
  \vspace{-0.1in}
  \hfil
  \subfloat[Hyundai Elantra Touring Hatchback 2012 from Stanford Cars]{\includegraphics[width=0.9\hsize]{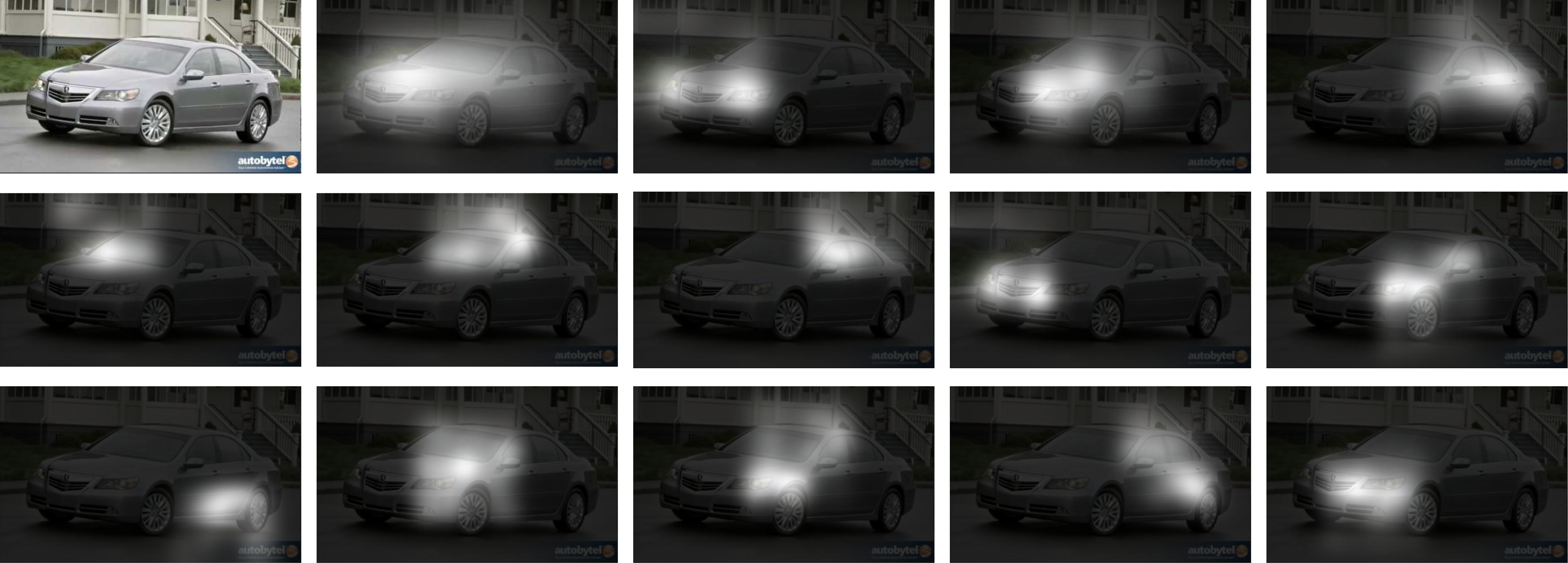}}
  \hfil
  \caption{Three images and their attention maps of different time steps from (a) CUB-200-2011 (b) Stanford Dogs (c) Stanford Cars.}
  \label{fig:visualization}
  \vspace{-0.2in}
\end{figure}
Fig.~\ref{fig:visualization} visualizes the attention maps generated by DVAN. To save the space, 15 out of 32 attention maps are selected for demonstration. The first attention map is generated using $224\times 224$ attention canvas, followed by six attention maps generated using $168\times 168$ attention canvases. The last eight attention maps are generated by the attention canvases with size of $112 \times 112$.

Generally, the attention maps generated by small scale attention canvases incorporate more areas of the target object, while the attention maps produced by large scale canvases enclose some parts of the object, having a higher resolution. We can see that the diversified attention maps are correctly detected across different time steps. In Fig.~\ref{fig:visualization}(a), the whole body of the bird is first observed, then the beak and neck areas are focused. In the last few time steps, local regions of tail and body receive more attention. Similarly, in Fig.~\ref{fig:visualization}(b), the main body of the dog is firstly attended. The head, body and legs are then attended sequentially using larger scale attention canvases. For Fig.~\ref{fig:visualization}(c), the frontal part of the car is first attended using the small scale attention canvas. After enlarging the local parts, the hoods, bumpers, and side doors are observed. Finally, the front bumps and wheels are attended as well. However, our experiments also indicate that tiny local parts of cars will reduce the classification accuracy, since these parts dismiss the ability to distinguish the models of cars.

\section{Conclusion}
\label{sec:conclusion}

In this paper, a diversified visual attention network is proposed for fine-grained object classification, which explicitly diversifies the attention maps for localizing multiple discriminative parts. By using a recurrent soft attention mechanism, the proposed framework dynamically attends important regions at each time step. Experiments on three publicly available datasets demonstrate that DVAN achieves competitive performance with state-of-the-art approaches, without using any information of bounding box or part location. In the future, we will explore the impact of attributes in fine-grained object classification problems, in which the attribute information will be exploited to guide the attention model for better discovering the discriminative regions.

\bibliographystyle{IEEEtran}
\bibliography{TMM_DVAN}

\begin{thebibliography}{10}
\providecommand{\url}[1]{#1}
\csname url@samestyle\endcsname
\providecommand{\newblock}{\relax}
\providecommand{\bibinfo}[2]{#2}
\providecommand{\BIBentrySTDinterwordspacing}{\spaceskip=0pt\relax}
\providecommand{\BIBentryALTinterwordstretchfactor}{4}
\providecommand{\BIBentryALTinterwordspacing}{\spaceskip=\fontdimen2\font plus
\BIBentryALTinterwordstretchfactor\fontdimen3\font minus
  \fontdimen4\font\relax}
\providecommand{\BIBforeignlanguage}[2]{{%
\expandafter\ifx\csname l@#1\endcsname\relax
\typeout{** WARNING: IEEEtran.bst: No hyphenation pattern has been}%
\typeout{** loaded for the language `#1'. Using the pattern for}%
\typeout{** the default language instead.}%
\else
\language=\csname l@#1\endcsname
\fi
#2}}
\providecommand{\BIBdecl}{\relax}
\BIBdecl

\bibitem{CUB_200_2011}
C.~Wah, S.~Branson, P.~Welinder, P.~Perona, and S.~Belongie, ``{The
  Caltech-UCSD Birds-200-2011 Dataset},'' 2011.

\bibitem{standford_dogs}
A.~Khosla, N.~Jayadevaprakash, B.~Yao, and L.~Fei-Fei, ``Novel dataset for
  fine-grained image categorization,'' in \emph{Proc. IEEE Int. Conf. Comput.
  Vis. Pattern Recog. Workshops}, Jun. 2011, pp. 3466--3473.

\bibitem{Liu2012}
J.~Liu, A.~Kanazawa, D.~Jacobs, and P.~Belhumeur, ``Dog breed classification
  using part localization,'' in \emph{Proc. IEEE Euro. Conf. Comput. Vis.},
  Oct. 2012, pp. 172--185.

\bibitem{standford_cars}
J.~Krause, M.~Stark, J.~Deng, and L.~Fei-Fei, ``3d object representations for
  fine-grained categorization,'' in \emph{Proc. IEEE Int. Conf. Comput. Vis.
  Workshops}, Jun. 2013, pp. 554--561.

\bibitem{Yang:2015ul}
L.~Yang, P.~Luo, C.~Change~Loy, and X.~Tang, ``A large-scale car dataset for
  fine-grained categorization and verification,'' in \emph{Proc. IEEE Conf.
  Comput. Vis. Pattern Recog.}, Jun. 2015, pp. 3973--3981.

\bibitem{tmm04}
L.~Xie, J.~Wang, B.~Zhang, and Q.~Tian, ``Fine-grained image search,''
  \emph{IEEE Trans. Multimedia}, vol.~17, no.~5, pp. 636--647, May 2015.

\bibitem{r03}
K.~Yamaguchi, M.~H. Kiapour, L.~E. Ortiz, and T.~L. Berg, ``Retrieving similar
  styles to parse clothing,'' \emph{IEEE Trans. Pattern Anal. and Mach.
  Intell.}, vol.~37, no.~5, pp. 1028--1040, May 2015.

\bibitem{r06}
X.~Wang, T.~Zhang, D.~R. Tretter, and Q.~Lin, ``Personal clothing retrieval on
  photo collections by color and attributes,'' \emph{IEEE Trans Multimedia},
  vol.~15, no.~8, pp. 2035--2045, Dec. 2013.

\bibitem{r08}
X.~Liang, L.~Lin, W.~Yang, P.~Luo, J.~Huang, and S.~Yan, ``Clothes co-parsing
  via joint image segmentation and labeling with application to clothing
  retrieval,'' \emph{IEEE Trans. Multimedia}, vol.~18, no.~6, pp. 1175--1186,
  Jun. 2016.

\bibitem{r09}
M.~Bola{\~{n}}os and P.~Radeva, ``Simultaneous food localization and
  recognition,'' in \emph{Proc. Int. Conf. Pattern Recog.}, 2016.

\bibitem{attention_categorization}
P.~Sermanet, A.~Frome, and E.~Real, ``Attention for fine-grained
  categorization,'' in \emph{Proc. Int. Conf. Learning Representations}, May
  2015.

\bibitem{tmm07}
L.~Zhu, J.~Shen, H.~Jin, L.~Xie, and R.~Zheng, ``Landmark classification with
  hierarchical multi-modal exemplar feature,'' \emph{IEEE Trans. Multimedia},
  vol.~17, no.~7, pp. 981--993, Jul. 2015.

\bibitem{wei2015hcp}
Y.~Wei, W.~Xia, M.~Lin, J.~Huang, B.~Ni, J.~Dong, Y.~Zhao, and S.~Yan, ``Hcp: A
  flexible cnn framework for multi-label image classification,'' \emph{IEEE
  Trans. Pattern Recog. and Mach. Intell.}, vol.~38, no.~9, pp. 1901--1907,
  2016.

\bibitem{poof}
T.~Berg and P.~N. Belhumeur, ``{POOF}: {P}art-{B}ased {O}ne-vs-{O}ne {F}eatures
  for fine-grained categorization, face verification, and attribute
  estimation,'' in \emph{Proc. IEEE Conf. Comput. Vis. Pattern Recog.}, 2013,
  pp. 955--962.

\bibitem{kernal}
L.~Bo, X.~Ren, and D.~Fox, ``Kernel descriptors for visual recognition,'' in
  \emph{Proc. Adv. Neural Inf. Process. Syst.}, Dec. 2010, pp. 244--252.

\bibitem{pose_normalized}
S.~Branson, G.~V. Horn, S.~Belongie, and P.~Perona, ``Bird species
  categorization using pose normalized deep convolutional nets,'' in
  \emph{Proc. British Mach. Vis. Conf.}, 2014, pp. 1--14.

\bibitem{symbiotic_segmentation}
Y.~Chai, V.~Lempitsky, and A.~Zisserman, ``Symbiotic segmentation and part
  localization for fine-grained categorization,'' in \emph{Proc. IEEE Int.
  Conf. Comput. Vis.}, Dec. 2013, pp. 321--328.

\bibitem{codebook_free}
B.~Yao, ``A codebook-free and annotation-free approach for fine-grained image
  categorization,'' in \emph{Proc. IEEE Conf. Comput. Vis. Pattern Recog.},
  2012, pp. 3466--3473.

\bibitem{part-based-rcnn}
N.~Zhang, J.~Donahue, R.~B. Girshick, and T.~Darrell, ``Part-based r-cnns for
  fine-grained category detection,'' in \emph{Proc. IEEE Euro. Conf. Comput.
  Vis.}, 2014, pp. 834--849.

\bibitem{two_level_attention}
T.~Xiao, Y.~Xu, K.~Yang, J.~Zhang, Y.~Peng, and Z.~Zhang, ``The application of
  two-level attention models in deep convolutional neural network for
  fine-grained image classification,'' in \emph{Proc. IEEE Conf. Comput. Vis.
  Pattern Recog.}, 2015, pp. 842--850.

\bibitem{fully_convolutional_attention}
X.~Liu, T.~Xia, J.~Wang, and Y.~Lin, ``Fully convolutional attention
  localization networks: Efficient attention localization for fine-grained
  recognition,'' \emph{ArXiv:abs/1603.06765}, 2016.

\bibitem{ronald_2000}
R.~A. Rensink, ``The dynamic representation of scenes,'' \emph{Visual
  Cognition}, vol.~7, no. 1-3, pp. 17--42, 2000.

\bibitem{tmm01}
T.~V. Nguyen, B.~Ni, H.~Liu, W.~Xia, J.~Luo, M.~Kankanhalli, and S.~Yan,
  ``Image re-attentionizing,'' \emph{IEEE Trans. Multimedia}, vol.~15, no.~8,
  pp. 1910--1919, Dec. 2013.

\bibitem{action_recognition}
S.~Sharma, R.~Kiros, and R.~Salakhutdinov, ``Action recognition using visual
  attention,'' in \emph{NIPS Time Series Workshop}, Dec. 2015.

\bibitem{multiple-object_visual-attention}
J.~Ba, V.~Mnih, and K.~Kavukcuoglu, ``Multiple object recognition with visual
  attention,'' in \emph{Proc. Int. Conf. Learning Representations}, May 2015,
  pp. 1--10.

\bibitem{Xu:2015ut}
K.~Xu, J.~Ba, R.~Kiros, K.~Cho, A.~C. Courville, R.~Salakhutdinov, R.~S. Zemel,
  and Y.~Bengio, ``Show, attend and tell: Neural image caption generation with
  visual attention,'' in \emph{Int. Conf. Mach. Learning}, 2015, pp.
  2048--2057.

\bibitem{Jaderberg:2015vo}
M.~Jaderberg, K.~Simonyan, A.~Zisserman, and k.~kavukcuoglu, ``Spatial
  transformer networks,'' in \emph{Adv. Neural Inf. Process. Syst.}, 2015, pp.
  2017--2025.

\bibitem{Hochreiter:LSM}
S.~Hochreiter and J.~Schmidhuber, ``Long short-term memory,'' \emph{Neural
  Comput.}, vol.~9, no.~8, pp. 1735--1780, Nov. 1997.

\bibitem{alexnet}
A.~Krizhevsky, I.~Sutskever, and G.~E. Hinton, ``Imagenet classification with
  deep convolutional neural networks,'' in \emph{Proc. Adv. Neural Inf.
  Process. Syst.}, 2012, pp. 1097--1105.

\bibitem{decaf}
J.~Donahue, Y.~Jia, O.~Vinyals, J.~Hoffman, N.~Zhang, E.~Tzeng, and T.~Darrell,
  ``Decaf: A deep convolutional activation feature for generic visual
  recognition,'' in \emph{Proc. Int. Conf. Mach. Learning}, 2014, pp. 647--655.

\bibitem{Simonyan:2014ws}
K.~Simonyan and A.~Zisserman, ``Very deep convolutional networks for
  large-scale image recognition,'' in \emph{Proc. Int. Conf. Learning
  Representations}, May 2015.

\bibitem{bilinear}
T.-Y. Lin, A.~RoyChowdhury, and S.~Maji, ``Bilinear cnn models for fine-grained
  visual recognition,'' in \emph{Proc. IEEE Int. Conf. Comput. Vis.}, 2015, pp.
  1449--1457.

\bibitem{Wang:2015vo}
Z.~Wang, X.~Wang, and G.~Wang, ``Learning fine-grained features via a {CNN}
  tree for large-scale classification,'' \emph{ArXiv:abs/1511.04534}, 2015.

\bibitem{tmm06}
C.~Luo, B.~Ni, S.~Yan, and M.~Wang, ``Image classification by selective
  regularized subspace learning,'' \emph{IEEE Trans. Multimedia}, vol.~18,
  no.~1, pp. 40--50, Jan. 2016.

\bibitem{Ge:2015tr}
Z.~Ge, C.~McCool, C.~Sanderson, and P.~I. Corke, ``Subset feature learning for
  fine-grained category classification,'' in \emph{Proc. IEEE Conf. Comput.
  Vis. Pattern Recog.}, 2015, pp. 46--52.

\bibitem{template}
S.~Yang, L.~Bo, J.~Wang, and L.~G. Shapiro, ``Unsupervised template learning
  for fine-grained object recognition,'' in \emph{Adv. Neural Inf. Process.
  Syst.}, 2012, pp. 3122--3130.

\bibitem{alignment}
E.~Gavves, B.~Fernando, C.~G.~M. Snoek, A.~W.~M. Smeulders, and T.~Tuytelaars,
  ``Fine-grained categorization by alignments,'' in \emph{Proc. IEEE Int. Conf.
  Comput. Vis.}, Dec. 2013, pp. 1713--1720.

\bibitem{Chai11}
Y.~Chai, V.~Lempitsky, and A.~Zisserman, ``Bicos: A bi-level co-segmentation
  method for image classification,'' in \emph{Proc. IEEE Int. Conf. Comput.
  Vis.}, Nov. 2011, pp. 2579--2586.

\bibitem{improved_pose_normalized}
S.~Branson, G.~Van~Horn, P.~Perona, and S.~J. Belongie, ``Improved bird species
  recognition using pose normalized deep convolutional nets.'' in \emph{Proc.
  British Mach. Vis. Conf.}, Sep. 2014, pp. 1--14.

\bibitem{Branson2010}
S.~Branson, C.~Wah, F.~Schroff, B.~Babenko, P.~Welinder, P.~Perona, and
  S.~Belongie, ``Visual recognition with humans in the loop,'' in \emph{Proc.
  IEEE Euro. Conf. Comput. Vis.}, 2010, pp. 438--451.

\bibitem{crowdsourcing_fine_grained}
J.~Deng, J.~Krause, and L.~Fei-Fei, ``Fine-grained crowdsourcing for
  fine-grained recognition,'' in \emph{Proc. IEEE Conf. Comput. Vis. Pattern
  Recog.}, Jun. 2013, pp. 580--587.

\bibitem{wah:similarity}
C.~Wah, G.~V. Horn, S.~Branson, S.~Maji, P.~Perona, and S.~Belongie,
  ``Similarity comparisons for interactive fine-grained categorization,'' in
  \emph{Proc. IEEE Conf. Comput. Vis. Pattern Recog.}, Jun. 2014, pp. 859--866.

\bibitem{ram}
V.~Mnih, N.~Heess, A.~Graves, and k.~kavukcuoglu, ``Recurrent models of visual
  attention,'' in \emph{Proc. Adv. Neural Inf. Process. Syst.}, 2014, pp.
  2204--2212.

\bibitem{reinforcement}
R.~J. Williams, ``Simple statistical gradient-following algorithms for
  connectionist reinforcement learning,'' \emph{Mach. Learn.}, vol.~8, no. 3-4,
  pp. 229--256, May 1992.

\bibitem{lstm}
W.~Zaremba, I.~Sutskever, and O.~Vinyals, ``Recurrent neural network
  regularization,'' \emph{ArXiv:abs/1409.2329}, 2014.

\bibitem{nac}
M.~Simon and E.~Rodner, ``Neural activation constellations: Unsupervised part
  model discovery with convolutional networks,'' in \emph{Proc. IEEE Int. Conf.
  Comput. Vis.}, 2015.

\bibitem{Zhang_2016_CVPR}
X.~Zhang, H.~Xiong, W.~Zhou, W.~Lin, and Q.~Tian, ``Picking deep filter
  responses for fine-grained image recognition,'' in \emph{Proc. IEEE Conf.
  Comput. Vis. Pattern Recog.}, June 2016, pp. 1134--1142.

\bibitem{bn}
S.~Ioffe and C.~Szegedy, ``Batch normalization: Accelerating deep network
  training by reducing internal covariate shift,'' in \emph{Proc. Int. Conf.
  Mach. Learning}, 2015, pp. 448--456.

\bibitem{Chatfield14}
K.~Chatfield, K.~Simonyan, A.~Vedaldi, and A.~Zisserman, ``Return of the devil
  in the details: Delving deep into convolutional nets,'' in \emph{British
  Mach. Vis. Conf.}, 2014.

\bibitem{selective}
J.~R. Uijlings, K.~E. van~de Sande, T.~Gevers, and A.~W. Smeulders, ``Selective
  search for object recognition,'' \emph{Int. J. Comput. Vis.}, vol. 104,
  no.~2, pp. 154--171, 2013.

\bibitem{Gavves2015}
E.~Gavves, B.~Fernando, C.~G.~M. Snoek, A.~W.~M. Smeulders, and T.~Tuytelaars,
  ``Local alignments for fine-grained categorization,'' \emph{Int. J. of
  Comput. Vis.}, vol. 111, no.~2, pp. 191--212, 2015.

\bibitem{noisy}
J.~Krause, B.~Sapp, A.~Howard, H.~Zhou, A.~Toshev, T.~Duerig, J.~Philbin, and
  F.~Li, ``The unreasonable effectiveness of noisy data for fine-grained
  recognition,'' in \emph{Proc. IEEE Euro. Conf. Comput. Vis.}, Oct. 2016, pp.
  301--320.

\bibitem{ellf}
J.~Krause, T.~Gebru, J.~Deng, L.~J. Li, and L.~Fei-Fei, ``Learning features and
  parts for fine-grained recognition,'' in \emph{Proc. Int. Conf. Pattern
  Recog.}, Aug. 2014, pp. 26--33.

\bibitem{revisit_fv}
P.-H. Gosselin, N.~Murray, H.~Jégou, and F.~Perronnin, ``Revisiting the fisher
  vector for fine-grained classification,'' \emph{Pattern Recog. Letters},
  vol.~49, pp. 92--98, 2014.

\bibitem{rich_feature}
R.~Girshick, J.~Donahue, T.~Darrell, and J.~Malik, ``Rich feature hierarchies
  for accurate object detection and semantic segmentation,'' in \emph{Proc.
  IEEE Conf. Comput. Vis. Pattern Recog.}, Sep. 2014, pp. 580--587.

\end{thebibliography}

\begin{IEEEbiography}
[{\includegraphics[width=1in,height=1.25in,clip,keepaspectratio]{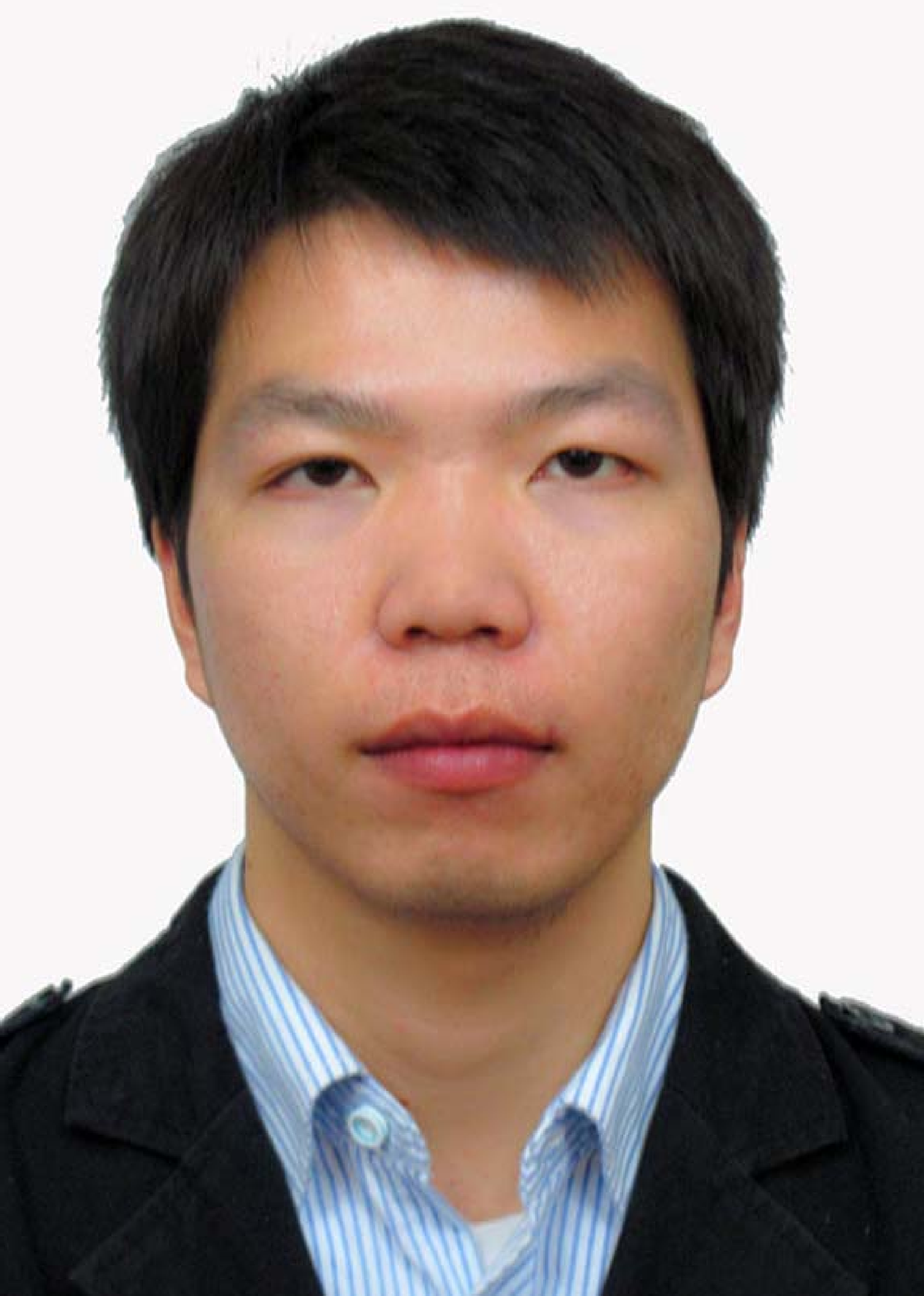}}]{Bo Zhao} is pursuing his Ph.D. degree from School of Information Science and Technology, Southwest Jiaotong University, Chengdu, China. Currently, he is at the Department of Electrical and Computer Engineering, National University of Singapore, Singapore as a Visiting Scholar. He received the B.Sc. degree in Networking Engineering from Southwest Jiaotong University in 2010. His research interests include multimedia, computer vision and machine learning.
\end{IEEEbiography}

\begin{IEEEbiography}
[{\includegraphics[width=1in,height=1.25in,clip,keepaspectratio]{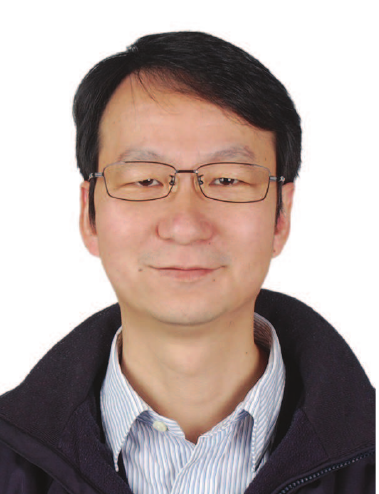}}]{Xiao Wu}
	(S'05-M'08) received the B.Eng. and M.S. degrees in computer science from Yunnan University, Yunnan, China, in 1999 and 2002, respectively, and the Ph.D. degree in Computer Science from City University of Hong Kong, Kowloon, in 2008.

  Currently, he is a Professor at Southwest Jiaotong University, Chengdu, China. He is the Assistant Dean of School of Information Science and Technology, and the Head of Department of Computer Science and Technology. He was with the Institute of Software, Chinese Academy of Sciences, Beijing, China, from 2001 to 2002. He was a Research Assistant and a Senior Research Associate at the City University of Hong Kong from 2003 to 2004, and 2007 to 2009, respectively. From 2006 to 2007, he was with the School of Computer Science, Carnegie Mellon University, Pittsburgh, PA, USA, as a Visiting Scholar, and at School of Information and Computer Science, University of California, Irvine, CA, USA as a Visiting Associate Professor during 2015 to 2016. He received the second prize of Natural Science Award of the Ministry of Education, China in 2015. His research interests include multimedia information retrieval, image/video computing, and data mining.
\end{IEEEbiography}

\begin{IEEEbiography}
[{\includegraphics[width=1in,height=1.25in,clip,keepaspectratio]{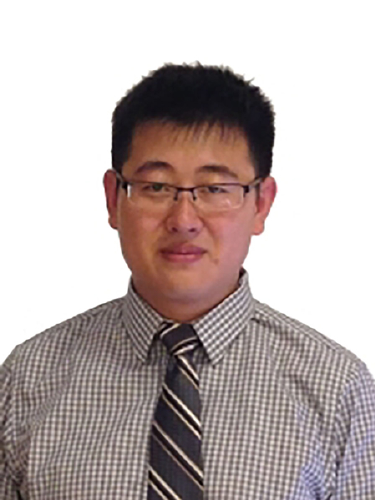}}]{Jiashi Feng} is currently an Assistant Professor in the Department of Electrical and Computer Engineering at the National University of Singapore. He got his B.E. degree from University of Science and Technology, China in 2007 and Ph.D. degree from National University of Singapore in 2014. He was a postdoc researcher at University of California from 2014 to 2015. His current research interest focuses on machine learning and computer vision techniques for large-scale data analysis. Specifically, he has done work in object recognition, deep learning, machine learning, high-dimensional statistics and big data analysis.
\end{IEEEbiography}

\begin{IEEEbiography}
[{\includegraphics[width=1in,height=1.25in,clip,keepaspectratio]{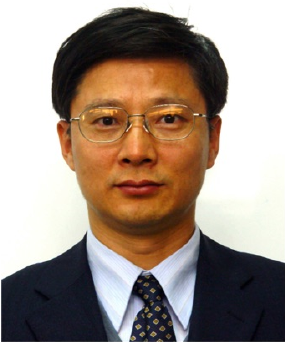}}]{Qiang Peng}
	received the B.E in automation control from Xi'an Jiaotong University, Xi'an, China, the M.Eng in computer application and technology, and the Ph.D. degree in traffic information and control engineering from Southwest Jiaotong University, Chengdu, China, in 1984, 1987, and 2004, respectively. He is currently a Professor at the School of Information Science and Technology, Southwest Jiaotong University, Chengdu, China. He has been in charge of more than 10 national scientific projects, published over 80 papers and holds 10 Chinese patents. His research interests include digital video compression and transmission, image/graphics processing, traffic information detection and simulation, virtual reality technology, multimedia system and application.
\end{IEEEbiography}

\begin{IEEEbiography}
[{\includegraphics[width=1in,height=1.25in,clip,keepaspectratio]{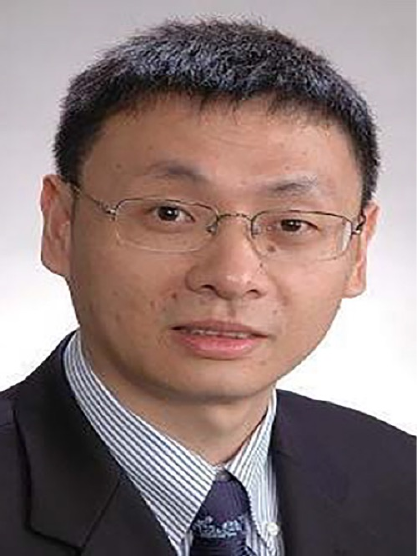}}]{Shuicheng Yan}
	(M'06-SM'09-F'17) is currently an Associate Professor at the Department National University of Singapore, and the founding lead of the Learning and Vision Research Group (http://www.lv-nus.org). Dr. Yan's research areas include machine learning, computer vision and multimedia, and he has authored/co-authored nearly 400 technical papers over a wide range of research topics, with Google Scholar citation$>$12,000 times. He is ISI highly-cited researcher 2014, and IAPR Fellow 2014. He has been serving as an associate editor of IEEE TKDE, CVIU and TCSVT. He received the Best Paper Awards from ACM MM'13 (Best Paper and Best Student Paper), ACM MM'12 (Best Demo), PCM'11, ACM MM'10, ICME'10 and ICIMCS'09, the runnerup prize of ILSVRC'13, the winner prizes of the classification task in PASCAL VOC 2010-2012, the winner prize of the segmentation task in PASCAL VOC 2012, the honorable mention prize of the detection task in PASCAL VOC'10, 2010 TCSVT Best Associate Editor (BAE) Award, 2010 Young Faculty Research Award, 2011 Singapore Young Scientist Award, and 2012 NUS Young Researcher Award.
\end{IEEEbiography}

\end{document}